%% file: main.tex
\pdfoutput=1

\documentclass[10pt,twocolumn,letterpaper]{article}

\usepackage[pagenumbers]{cvpr}

\input{preamble}

\usepackage[pagebackref,breaklinks,colorlinks,allcolors=accent]{hyperref}
\title{RL Post-Training in Diffusion Transformers Is Activation Steering}

\iflatexml
\author{Tuna Han Salih Meral$^{1}$, Kaan Oktay$^{2}$, Alec Helbling$^{3,4}$, Atakan Kara$^{2}$, Duen Horng Chau$^{3}$, Pinar Yanardag$^{1}$ \\ $^{1}$Virginia Tech, $^{2}$fal, $^{3}$Georgia Tech, $^{4}$IBM Research}
\else
\author{Tuna Han Salih Meral$^{1}$, Kaan Oktay$^{2}$, Alec Helbling$^{3,4}$, Atakan Kara$^{2}$, Duen Horng Chau$^{3}$, Pinar Yanardag$^{1}$}
\affiliations{$^{1}$Virginia Tech, $^{2}$fal, $^{3}$Georgia Tech, $^{4}$IBM Research}
\fi
\projectpage{https://rl-is-steering.github.io}

\begin{document}
\input{sec/0_abstract}
\maketitle
\begin{strip}
\centering
\includegraphics[width=\linewidth]{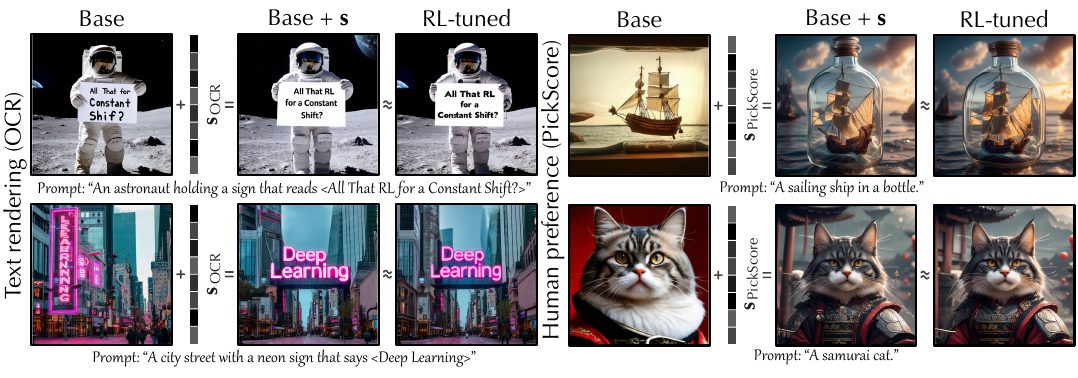}
\captionsetup{hypcap=false}
\captionof{figure}{\textbf{A constant shift of a frozen base model recovers 90.0\% to 96.3\% of the RL reward gain on six image models.}
\textbf{Base + $\mathbf{s}$} adds the same steering vectors, $\mathbf{s}_{\mathrm{OCR}}$ or $\mathbf{s}_{\mathrm{PickScore}}$, to the frozen \textbf{Base} SD3.5-M for every prompt, token and seed.
Each set (0.29M parameters, against 18.8M in the rank-32 LoRA) is fitted to its \textbf{RL-tuned} model with no reward in the objective.
Each triplet shares one prompt and starting noise.}
\captionsetup{hypcap=true}
\label{fig:teaser}%
\end{strip}
\begin{figure*}[t]
\centering
\includegraphics{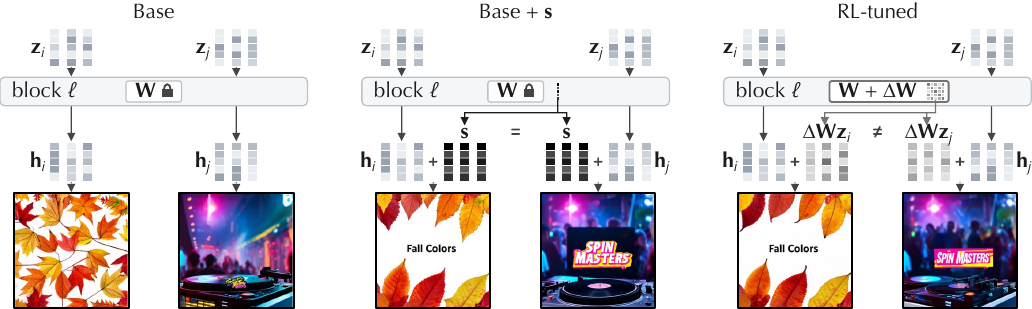}
\caption{\textbf{The steering vectors add the same vector to every token of every prompt.}
Block~$\ell$ on two text-rendering prompts, with a linear layer $\mathbf{W}$, token inputs $\mathbf{z}_i$ and $\mathbf{z}_j$, and block outputs $\mathbf{h}_i$ and $\mathbf{h}_j$.
\textbf{Base + $\mathbf{s}$} adds the same $\mathbf{s}$ (black) to every token of both prompts; $\mathbf{s}$ changes only with the block, guidance pass, step bin and stream.
In the \textbf{RL-tuned} model, the weight change gives each input its own shift, $\Delta\mathbf{W}\mathbf{z}_i \neq \Delta\mathbf{W}\mathbf{z}_j$ (gray). In block~$\ell$, $\mathbf{s}$ is one vector and $\Delta\mathbf{W}$ a matrix: on SD3.5-M, all 188 vectors together hold 0.29M parameters, 65$\times$ fewer than the RL-tuned LoRA's 18.8M.}
\label{fig:shared-shift}
\end{figure*}
\input{sec/1_intro}

\input{sec/2_related}
\input{sec/3_method}
\input{sec/4_experiments}

\input{sec/5_analysis}

\input{sec/6_conclusion}

{
    \small
    \bibliographystyle{ieeenat_fullname}
    \bibliography{main}
}

\input{sec/X_suppl}

\end{document}

%% file: preamble.tex
\usepackage{currfile} %

\usepackage{caption} %
\usepackage{microtype}

\newtheorem{theorem}{Theorem}

\expandafter\ifx\csname iflatexml\endcsname\relax\expandafter\newif\csname iflatexml\endcsname\fi
\iflatexml
\definecolor{accent}{HTML}{861F41}
\newcommand{\headfont}{}
\newcommand{\affiliations}[1]{}
\newcommand{\projectpage}[1]{}
\newenvironment{strip}{}{}
\else
\usepackage{cuted}
\usepackage{tcolorbox}
\definecolor{titlebox}{HTML}{F4F1EE}
\definecolor{accent}{HTML}{861F41}
\pdfmapline{=SourceSans3-Bold-tlf-ot1 SourceSans3-Bold "AutoEnc_gjrb6nmy77onppa5rlasmreh5d ReEncodeFont" <[a_gjrb6n.enc <SourceSans3-Bold.pfb}
\pdfmapline{=SourceSans3-Regular-tlf-ot1 SourceSans3-Regular "AutoEnc_gjrb6nmy77onppa5rlasmreh5d ReEncodeFont" <[a_gjrb6n.enc <SourceSans3-Regular.pfb}
\DeclareFontFamily{OT1}{arxivsans}{}
\DeclareFontShape{OT1}{arxivsans}{m}{n}{<-> SourceSans3-Regular-tlf-ot1}{}
\DeclareFontShape{OT1}{arxivsans}{b}{n}{<-> SourceSans3-Bold-tlf-ot1}{}
\DeclareFontShape{OT1}{arxivsans}{bx}{n}{<-> ssub * arxivsans/b/n}{}
\newcommand{\headfont}{\fontencoding{OT1}\fontfamily{arxivsans}\selectfont}

\makeatletter
\newcommand{\affiliations}[1]{\gdef\arxiv@affiliations{#1}}
\newcommand{\projectpage}[1]{\gdef\arxiv@projectpage{#1}}
\RenewDocumentEnvironment{abstract}{+b}{\gdef\arxiv@abstract{#1}}{}
\def\@maketitle{%
  \newpage\null
  \begin{tcolorbox}[colback=titlebox,colframe=titlebox,boxrule=0pt,arc=9pt,left=15pt,right=15pt,top=13pt,bottom=11pt]
    {\raggedright\headfont\bfseries\fontsize{19}{22}\selectfont\@title\par}
    \vskip 9pt
    {\raggedright\headfont\bfseries\normalsize\@author\par}
    \vskip 5pt
    {\raggedright\normalsize\arxiv@affiliations\par}
    \vskip 11pt
    {\normalsize\noindent\ignorespaces\arxiv@abstract\par}
    \vskip 9pt
    {\noindent\small\begin{minipage}[c]{0.51\linewidth}\raggedright{\headfont\bfseries Project page:} \url{\arxiv@projectpage}\end{minipage}\hfill
     \begin{minipage}[c]{0.47\linewidth}\raggedleft
       \raisebox{-0.5\height}{\includegraphics[height=0.33in]{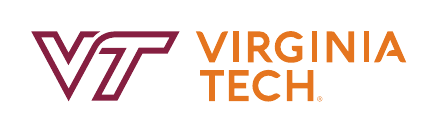}}\hspace{0.26in}\raisebox{-0.5\height}{\includegraphics[height=0.2in]{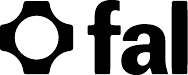}}\hspace{0.26in}\raisebox{-0.5\height}{\includegraphics[height=0.27in]{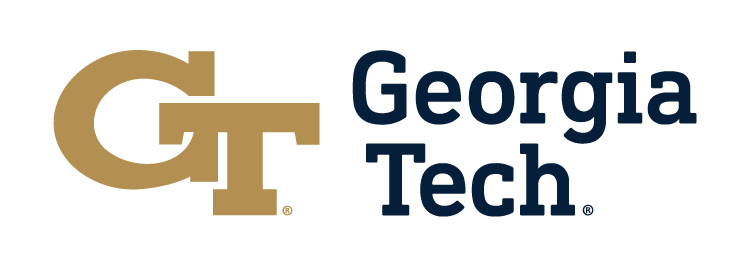}}
     \end{minipage}\par}
  \end{tcolorbox}
  \vskip 8pt
}
\def\cvprsection{\@startsection{section}{1}{\z@}{-12pt plus -2pt minus -2pt}{6pt}{\headfont\bfseries\large}}
\def\cvprsubsection{\@startsection{subsection}{2}{\z@}{-9pt plus -2pt minus -2pt}{4pt}{\headfont\bfseries\normalsize}}
\def\cvprsubsubsection{\@startsection{subsubsection}{3}{\z@}{-6pt plus -2pt minus -2pt}{3pt}{\headfont\bfseries\small}}
\makeatother
\DeclareCaptionFont{sansbold}{\renewcommand{\textbf}[1]{{\headfont\bfseries ##1}}}
\DeclareCaptionLabelFormat{sans}{{\headfont\bfseries #1 #2}}
\fi

%% file: sec/0_abstract.tex
\begin{abstract}
Reinforcement learning (RL) post-training substantially improves the visual quality, text rendering, and compositional accuracy of diffusion transformers. Because capabilities such as rendering requested text or composing multiple objects are inherently prompt-dependent, a natural hypothesis is that improving them requires learning complex, input-dependent transformations. To isolate how much of these gains an input-independent shift explains, we probe each RL-tuned model with a deliberately simple parameterization: a fixed set of steering vectors added to the block outputs of the frozen base model. On SD3.5-M, these vectors hold 65$\times$ fewer parameters than the RL-tuned LoRA (0.29M against 18.8M). We fit them to match the velocity predictions of public RL-tuned checkpoints, with no reward in the objective. Although the vectors are identical for every prompt, token, and sampling seed, they recover 90.0--96.3\% of the reward gains of six RL-tuned image models, and 89.8\% and 111.9\% on two RL-tuned video models. Initialized at zero and trained with online RL, the vectors alone exceed the reward gains of weight-based RL on text rendering (OCR) and human preference (PickScore) on SD3.5-M. A constant shift thus suffices for most of the gain of policy-gradient RL post-training: without new input-dependent transformations, it steers the pretrained generator toward what it can already compute.

\end{abstract}

%% file: sec/1_intro.tex
\section{Introduction}
\label{sec:intro}

Reinforcement learning (RL) post-training~\cite{black2024training,fan2023dpok,liu2026flow,xue2025dancegrpo,zheng2026diffusionnft} improves text rendering~\cite{liu2026flow}, compositional accuracy~\cite{ghosh2023geneval,huang2023t2i} and human preference scores~\cite{kirstain2023pick,wu2023human,ma2025hpsv3,xu2023imagereward} in image and video diffusion transformers~\cite{esser2024scaling,labs2024flux,wan2025wan}. These improvements are typically obtained by updating the model's weights, either through parameter-efficient adapters such as LoRA~\cite{hu2021lora} or through full fine-tuning~\cite{xue2025dancegrpo}. Although the update $\Delta\mathbf{W}$ is fixed after training, the updated weights $\mathbf{W} + \Delta\mathbf{W}$ act on each token's input, so different prompts and tokens produce different activation changes. This flexibility seems natural for visual generation, where the words to render and the objects to compose change with every prompt. \textit{But is this input dependence actually necessary to obtain the gains of RL post-training?}

Surprisingly, we find that most of the reward improvement of a diverse collection of RL-tuned image and video generators can be reproduced by freezing every pretrained weight and adding a small set of fixed steering vectors to the model's activations. We fit these vectors to match the velocity predictions of each RL-tuned model, with no reward in the objective; simply averaging the activation changes that RL induces recovers far less of the gain. For a given RL-tuned model, the same vectors are used for every prompt, every sampling seed, and every token (Fig.~\ref{fig:shared-shift}). In total, they have 1.5\% of the parameters of a rank-32 LoRA on SD3.5-M, and they do not depend on the requested text, objects, or composition. Yet they can make the frozen generator render requested phrases, satisfy compositional constraints and improve preference scores much as its RL-tuned counterpart does (Figs.~\ref{fig:teaser} and~\ref{fig:image-examples}). On six RL-tuned image models spanning SD3.5-M and FLUX.1-dev, multiple RL algorithms, LoRA and full fine-tuning, and rewards including PickScore, HPSv2.1, OCR, and GenEval, these fixed activation shifts recover 90.0--96.3\% of the RL reward gain. The result extends to video: on two RL-tuned video generators, the same intervention recovers 89.8\% and 111.9\% of the gain.

We next ask whether RL can learn such constant shifts directly from the reward in diffusion transformers. We freeze the generator and run Flow-GRPO~\cite{liu2026flow} with only the steering vectors trainable, starting from zero. The trained vectors reach 104.6\% and 103.6\% of the gains of the OCR-tuned and PickScore-tuned SD3.5-M models, so RL achieves both gains by learning only a small set of constant vectors, without updating any pretrained weight.

How can one constant intervention produce different improvements for different prompts? The steering vectors carry no prompt information, so the prompt-specific part of the change must come from the frozen generator. Inside the transformer, attention, normalization and MLP layers mix the constant vector with prompt-dependent representations, so its effect on the output can depend strongly on the prompt. On SD3.5-M, we trace how the shift reaches the image: through what image tokens read from text tokens in joint attention. When image tokens read the text keys and values of the unsteered base model instead, with everything else unchanged, most of the OCR and PickScore gains disappear. The same replacement removes most of the gains of the OCR-tuned and PickScore-tuned models, so the steering vectors and RL reach the image by the same route. The text stream, the conditional guidance pass and the early sampling steps carry most of the gain: vectors fitted to these parts alone, with 531$\times$ fewer parameters than the rank-32 LoRA, recover 89.5\% of the RL gain on OCR and 73.7\% on PickScore (\cref{sec:ana_credit}).

These results suggest a different view of RL post-training in visual generators: a constant shift suffices to steer generation toward most of the RL gain, and the frozen generator does the prompt-specific work. Although RL's weight updates produce input-dependent activation changes, most of their reward gain is reproduced without new input-dependent transformations beyond those the base model already computes. Our main contributions are:

\begin{itemize}[leftmargin=*,itemsep=1pt,topsep=2pt]
\item We show that fixed steering vectors, fitted with no reward in the objective, recover most of the gain of eight RL-tuned image and video models, including two full fine-tunes.
\item We find that matching the RL-tuned model's velocity recovers far more of the gain than adding its average activation change.
\item We train only the steering vectors of a frozen SD3.5-M with Flow-GRPO and exceed the reward gains of the RL-tuned OCR and PickScore models, with images rated on par with theirs in a user study.
\item We trace how a constant shift yields prompt-specific gains: on SD3.5-M, it reaches the image through what image tokens read from text, as RL's change does.
\end{itemize}

%% file: sec/2_related.tex
\section{Related work}
\label{sec:related}

\noindent \textbf{RL post-training of diffusion transformers.}
Reward fine-tuning raises a score of a diffusion model's samples by backpropagating the score through the sampler~\cite{clark2024directly,shen2025directly}, by learning from preference pairs~\cite{wallace2024diffusion} or with policy gradients~\cite{black2024training,fan2023dpok}.
Policy gradients~\cite{williams1992simple} need only one score per sample, and their group-relative form in Flow-GRPO~\cite{liu2026flow} and DanceGRPO~\cite{xue2025dancegrpo} now post-trains image and video diffusion transformers~\cite{esser2024scaling,labs2024flux,wan2025wan}.
The rewards include OCR accuracy~\cite{liu2026flow}, GenEval~\cite{ghosh2023geneval} and T2I-CompBench~\cite{huang2023t2i} scores, and preference models trained on human judgments~\cite{christiano2017deep,kirstain2023pick,wu2023human,xu2023imagereward,ma2025hpsv3}.
Other studies report side effects on color saturation~\cite{pan2026normguard} and output diversity~\cite{bao2026renft}.
We ask what RL changes inside the model to produce these gains.

\noindent \textbf{Where RL acts.}
Studies of RL in language models ask where its update sits.
RL fine-tuning updates a small subnetwork~\cite{mukherjee2026reinforcement}, reaches most of its gain with few trained parameters~\cite{morris2026learning} and keeps the model close to the base model~\cite{shenfeld2026rl}.
Alignment tuning mainly changes the stylistic or first tokens of each answer~\cite{lin2024unlocking,qi2025safety}, and RL raises the probability of answers that the base model already samples~\cite{yue2026does}.
Venhoff \etal~\cite{venhoff2025base} close most of the gap to RL-trained reasoning models with classifier-gated steering vectors: a classifier decides where to add each reasoning behavior.
Sinii \etal~\cite{sinii2025steering,sinii2025small} train steering vectors with RL on a math reward and match or nearly match full RL fine-tuning, and Vakada \etal~\cite{vakada2026label} do the same with label-free rewards.
We fit ours to a given RL-tuned model to measure how much of its gain a constant recovers, and also train them with the reward (\cref{sec:rl-steering}).
We study diffusion transformers, in which a reward gain is a prompt-specific change of a whole image or video.

\noindent \textbf{Activation steering.}
Activation steering adds a fixed vector to a model's hidden states, taken from contrasting prompts~\cite{turner2024activation,li2023inference,panickssery2023steering,zou2023representation} or optimized~\cite{subramani2022extracting,cao2024personalized}.
In image models, a fixed direction in deep features edits image content~\cite{upchurch2017deep}, and similar, often gated, additions steer diffusion models~\cite{kwon2022diffusion,gaintseva2026casteer} and diffusion transformers~\cite{konovalova2026shift}.
Task arithmetic~\cite{ilharco2022editing} adds the weight difference of a fine-tuned model, which is constant in weight space but changes the activations of each input differently.
In concurrent work, First-Order Steering~\cite{kunda2026first} turns such a weight update into steering vectors to first order.
Adila \etal~\cite{adila2026weight} find the block output, where our steering vectors sit, an expressive place for such shifts.
We instead fit steering vectors to the velocity predictions of a given RL-tuned model.

\noindent \textbf{Constants in a network.}
Many adaptation methods change a frozen model through parameters that are the same for every input.
At the input, prompt tuning~\cite{lester2021power} trains soft prompt tokens, textual inversion~\cite{gal2022image} fits a word for a new concept and ReNeg~\cite{li2025reneg} learns a negative embedding on a reward.
RaTE~\cite{chen2026mapreduce} fits one prompt token to a Flow-GRPO model by flow matching.
Inside the network, prefix tuning~\cite{li2021prefix} and P-Tuning v2~\cite{liu2022p} prepend learned keys and values to attention layers, and ReFT~\cite{wu2024reft} learns interventions on hidden representations.
BitFit~\cite{zaken2022bitfit} and (IA)$^3$~\cite{liu2022few} train biases and channel scales, DiffFit~\cite{xie2023difffit} tunes the biases and scales of a diffusion transformer and Calibri~\cite{tokhchukov2026calibri} tunes the block gates of diffusion transformers on a reward.
BitFit and DiffFit train biases with the task loss to adapt a model to new data, whereas we fit steering vectors at the block outputs to an existing RL-tuned model to measure how much of its gain they reproduce.

%% file: sec/3_method.tex
\section{Method}
\label{sec:method}

\subsection{Preliminaries}
\label{sec:method_prelim}

\noindent\textbf{Diffusion transformers.}
The base model, the RL-tuned model and the steered base model are diffusion transformers~\cite{peebles2023scalable} trained by flow matching~\cite{liu2022flow,lipman2023flow}.
Each predicts a velocity $\mathbf{v}$ at the noisy latent $\mathbf{x}_t = (1-t)\mathbf{x}_0 + t\boldsymbol{\epsilon}$, with the clean latent $\mathbf{x}_0$, the noise level $t \in [0,1]$ and Gaussian noise $\boldsymbol{\epsilon} \sim \mathcal{N}(\mathbf{0}, \mathbf{I})$.
Let $\mathbf{v}_\theta$ denote the base model, $\mathbf{v}_{\theta_{\mathrm{RL}}}$ the RL-tuned model and $\mathbf{v}_{\theta,\mathbf{s}}$ the steered base model.
The sampler takes $N$ sampling steps from $t = 1$ to $0$, one per discrete noise level.
We use $N = 40$ on SD3.5-M~\cite{esser2024scaling}, as in Flow-GRPO~\cite{liu2026flow}.

\noindent\textbf{Guidance passes.}
Classifier-free guidance (CFG)~\cite{ho2022classifier} combines the conditional pass for the prompt $y$ and the unconditional pass for the empty prompt $\varnothing$ as ${\mathbf{v}_{\varnothing} + w(\mathbf{v}_{\mathrm{c}} - \mathbf{v}_{\varnothing})}$, with CFG scale $w$.
FLUX.1-dev~\cite{labs2024flux} is guidance-distilled and runs a single pass per sampling step.

\noindent\textbf{Token streams.}
SD3.5-M processes image and text tokens in two streams.
In joint attention~\cite{esser2024scaling}, every token attends to both streams.

\subsection{Steering vectors}
\label{sec:method_shift}
\begin{figure}[t]
\centering
\includegraphics{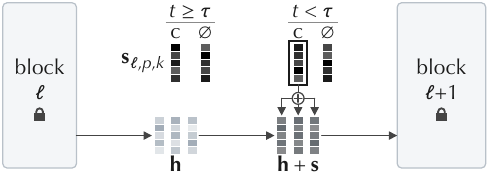}
\caption{\textbf{One steering vector shifts every token of a block output.}
The drawing shows one stream.
The guidance pass $p$ and step bin $k$ select the outlined vector, which is added to every token $\mathbf{h}$.}
\label{fig:shift-layout}
\end{figure}

A steering vector is a constant vector added to a model's hidden states at inference, which changes the model's output without changing its weights~\cite{subramani2022extracting,turner2024activation,panickssery2023steering}.
In a model of $L$ blocks and width $d$, let $\mathbf{h}^{\mathrm{str}}_{\ell,i} \in \mathbb{R}^d$ be token $i$ of the output of block $\ell \in \{0, \dots, L-1\}$ in the image or text stream, $\mathrm{str} \in \{\mathrm{img}, \mathrm{txt}\}$.
The steering vectors $\mathbf{s} = \{\mathbf{s}^{\mathrm{str}}_{\ell,p,k}\}$ add one vector in $\mathbb{R}^d$ to every token of each output (\cref{fig:shift-layout}):
\begin{equation}
  \tilde{\mathbf{h}}^{\mathrm{str}}_{\ell,i} = \mathbf{h}^{\mathrm{str}}_{\ell,i} + \alpha\,\mathbf{s}^{\mathrm{str}}_{\ell,p,k(t)}
  \quad \text{for every token } i.
  \label{eq:shift}
\end{equation}
The guidance pass $p \in \{\mathrm{c}, \varnothing\}$ and the step bin $k(t) \in \{1, \dots, K\}$ that contains $t$ select the vector, and the steering strength $\alpha$ scales it.
Fitting and all main results use $\alpha = 1$.
We use $K=2$ contiguous step bins with a fixed boundary $\tau$ that splits the schedule's $N$ noise levels equally between early ($t\geq\tau$) and late ($t<\tau$) steps, with $\tau = 0.75$ on SD3.5-M's shifted schedule.

Both $p$ and $k$ depend only on the sampler, so every prompt, sampling seed and token of a stream receives the same vectors.
A full or LoRA~\cite{hu2021lora} weight update $\Delta\mathbf{W}$ of a linear layer instead gives $(\mathbf{W} + \Delta\mathbf{W})\,\mathbf{z}_i = \mathbf{W}\mathbf{z}_i + \Delta\mathbf{W}\,\mathbf{z}_i$, with $\mathbf{z}_i$ the layer's input at token $i$ (\cref{fig:shared-shift}). The added term differs across tokens and prompts, unlike \cref{eq:shift}.
A rank-1 LoRA $\Delta\mathbf{W} = \mathbf{b}\mathbf{a}^\top$ adds the fixed direction $\mathbf{b}$ weighted by $\mathbf{a}^\top\mathbf{z}_i$, so it is a steering vector with an input-dependent weight.
On SD3.5-M, the steering vectors have 288,768 parameters (\cref{tab:shift-layouts}), 1.5\% of those in the rank-32 LoRA that Flow-GRPO~\cite{liu2026flow} trains for PickScore.
Single-pass models (FLUX.1-dev and Wan2.2-5B) drop the pass index.

\noindent\textbf{Shared across tokens.}
The steering vectors do not depend on the token count, so vectors fitted at one resolution apply unchanged at another.
They transfer from 512 to 768 and 1024\,px, and to other numbers of sampling steps and CFG scales (\cref{fig:fit-sensitivity}).

\noindent\textbf{Split by guidance pass.}
Guidance scales a change $\Delta\mathbf{v}_{\mathrm{c}}$ in the conditional velocity by $w$ and a change $\Delta\mathbf{v}_{\varnothing}$ in the unconditional velocity by $1-w$.
One vector shared by both passes cannot set the two changes independently, and the two factors have opposite signs whenever $w > 1$.

\noindent\textbf{Split by step bin.}
Early sampling steps set an image's layout, and late steps refine its detail~\cite{balaji2022ediffi,hertz2022prompt}.
The RL-tuned model's average block-output change also varies over the steps: two step bins keep 83\% to 94\% of the energy that eight keep (\cref{tab:step-bins}), and one bin gives nearly the same reward gain as two (\cref{tab:capacity-ladder}).

\begin{table*}[t]
\centering
\caption{\textbf{Steering vectors recover most of each RL-tuned model's reward gain.}
Gains are raw reward points over the base model.
Shares are of the RL-tuned model's gain, with the 95\% interval of the $\mathbf{s}_{\mathrm{fit}}$ share in brackets.}
\small
\input{tables/recovery-table}
\label{tab:recovery-table}
\end{table*}

\subsection{Fitting by velocity matching}
\label{sec:method_fit}

\noindent\textbf{Mean-difference baseline.} The simplest input-\hskip0pt independent change adds the mean-difference steering vectors $\mathbf{s}_{\mathrm{mean}}$, a training-free baseline in the layout of $\mathbf{s}$.
Each vector of $\mathbf{s}_{\mathrm{mean}}$ is the RL-tuned model's block output minus the base model's, averaged over tokens, prompts and the noise levels in its step bin.
Both models run on the same noisy states along the RL-tuned model's sampling trajectory.
We measure blocks $\ell = 0, \dots, L-1$ sequentially with earlier vectors already added to the base model, so no block's change is counted twice.
The vectors of $\mathbf{s}_{\mathrm{mean}}$ are the two-model counterpart of the contrastive mean-difference vector of activation steering~\cite{turner2024activation,panickssery2023steering}.

\begin{figure}[t]
\centering
\includegraphics[width=0.9032\linewidth]{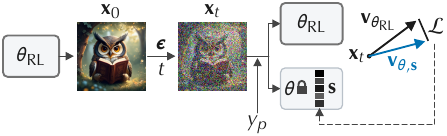}
\caption{\textbf{Fitting matches the RL-tuned model's velocity.}
The RL-tuned model's final latent $\mathbf{x}_0$ is re-noised to $\mathbf{x}_t$, and the loss $\mathcal{L}$ trains only $\mathbf{s}$ (dashed arrow), with no reward.}
\label{fig:shift-fitting}
\end{figure}

\noindent\textbf{Velocity matching.}
We instead fit the steering vectors to match the output velocity of the RL-tuned model, not its activation changes (\cref{fig:shift-fitting}), as RaTE~\cite{chen2026mapreduce} does for one prompt token.
Fitting minimizes
\begin{equation}
  \mathcal{L}(\mathbf{s}) = \mathbb{E}\,
  \big\lVert \mathbf{v}_{\theta,\mathbf{s}}(\mathbf{x}_t, t, y_p) - \mathbf{v}_{\theta_{\mathrm{RL}}}(\mathbf{x}_t, t, y_p) \big\rVert^2,
  \label{eq:fit}
\end{equation}
where the expectation runs over prompts $y$, the RL-tuned model's final samples $\mathbf{x}_0$ for $y$, and $t$, $\boldsymbol{\epsilon}$ and $p$, with $y_{\mathrm{c}} = y$, $y_{\varnothing} = \varnothing$ and $t$ drawn from the sampler's $N$ noise levels (\cref{sec:suppl_recipe}).
In $\mathbf{v}_{\theta,\mathbf{s}}$, \cref{eq:shift} adds only the vectors of pass $p$ and of the step bin that contains $t$, so \cref{eq:fit} splits into one independent term per pass and step bin.
Both models stay frozen, and minimizing \cref{eq:fit} gives the fitted steering vectors $\mathbf{s}_{\mathrm{fit}}$.
Velocity matching is a regression on cached samples of the RL-tuned model for training prompts, so the training loop needs neither a reward model nor sampling.
Building the cache, measuring $\mathbf{s}_{\mathrm{mean}}$ and fitting take at most 17 H100 GPU-hours per image model and 70 per video model (\cref{tab:hyperparameters}).
At test time, the steering vectors cost one vector addition per block output.

%% file: tables/recovery-table.tex
\setlength{\tabcolsep}{2.8pt}
\begin{tabular}{@{}llllrrrrrr@{\,}l@{}}
\toprule
 & & & & Prompts & \multicolumn{3}{c}{Gain over base} & \multicolumn{3}{c}{Share (\%)} \\
\cmidrule(lr){6-8}\cmidrule(l){9-11}
Base & RL method & Fine-tuning & Reward & $\times$ seeds & $\mathbf{s}_{\mathrm{mean}}$ & $\mathbf{s}_{\mathrm{fit}}$ & RL-tuned & $\mathbf{s}_{\mathrm{mean}}$ & \multicolumn{2}{c}{$\mathbf{s}_{\mathrm{fit}}$} \\
\midrule
SD3.5-M & Flow-GRPO & LoRA & PickScore & 2,048 $\times$ 1 & +0.952 & +1.416 & +1.560 & 61.0 & 90.8 & [89.8, 91.7] \\
 & Flow-GRPO, no KL & LoRA & PickScore & 2,048 $\times$ 1 & +1.198 & +1.612 & +1.693 & 70.8 & 95.2 & [94.4, 96.1] \\
 & Flow-GRPO & LoRA & OCR & 1,018 $\times$ 1 & +0.074 & +0.325 & +0.344 & 21.5 & 94.4 & [90.8, 97.9] \\
 & Flow-GRPO & LoRA & GenEval & 553 $\times$ 4 & +0.033 & +0.260 & +0.271 & 12.2 & 95.9 & [92.0, 99.7] \\
\midrule
FLUX.1-dev & Flow-GRPO & LoRA & PickScore & 2,048 $\times$ 1 & +0.985 & +1.370 & +1.522 & 64.7 & 90.0 & [88.9, 91.1] \\
 & DanceGRPO & Full & HPSv2.1 & 2,048 $\times$ 1 & +0.042 & +0.052 & +0.054 & 79.0 & 96.3 & [95.5, 97.2] \\
\midrule
Wan2.1-1.3B & GenRL GRPO & LoRA & HPSv3 & 256 $\times$ 2 & +0.777 & +1.465 & +1.309 & 59.3 & 111.9 & [106.5, 118.0] \\
\midrule
Wan2.2-5B & VBVR-Pro RLVR & Full & Rule score & 250 $\times$ 4 & +0.049 & +0.072 & +0.080 & 60.8 & 89.8 & [76.2, 105.6] \\
\bottomrule
\end{tabular}

%% file: sec/4_experiments.tex
\section{A constant shift suffices for most  RL gain}
\label{sec:experiments}

\subsection{Setup}
\label{sec:exp_setup}
\Cref{tab:recovery-table} lists the eight RL-tuned models, all public checkpoints released by their authors.
On SD3.5-M, Flow-GRPO~\cite{liu2026flow} gives the OCR-tuned and GenEval-tuned models and the PickScore-tuned model with and without the KL penalty.
On FLUX.1-dev~\cite{labs2024flux}, Flow-GRPO tunes a LoRA for PickScore, and DanceGRPO~\cite{xue2025dancegrpo} fine-tunes all weights for HPSv2.1~\cite{wu2023human}.
The Wan2.1 RL-tuned model is a GenRL~\cite{genrl} LoRA on Wan2.1-T2V-1.3B~\cite{wan2025wan}, trained with the LongCat-Video recipe~\cite{team2025longcat} on HPSv3~\cite{ma2025hpsv3} and VideoAlign~\cite{liu2026improving} rewards.
The VBVR-Pro RL-tuned model fine-tunes all weights of its base model VBVR-Pro~\cite{xu2026vbvr}, a supervised fine-tune of Wan2.2-TI2V-5B, by RL with a verifiable rule-based reward (RLVR).
Of the six public RL-tuned video models we evaluated, only these two gain on the evaluated reward with a 95\% interval above zero (\cref{tab:video-detail}).
We evaluate each SD3.5-M RL-tuned model on the test set of its reward, the two FLUX.1-dev RL-tuned models on the PickScore test prompts, the Wan2.1 RL-tuned model on VBench~\cite{huang2024vbench} test prompts and the VBVR-Pro RL-tuned model on in-domain test instances that the fit never uses, from its RL training tasks (\cref{tab:recovery-table}).
On FLUX.1-dev, the steering vectors are fitted on PickScore training prompts, so the PickScore test set holds out prompts from the same source.
A gain is a model's mean raw reward minus the base model's, from the same prompts and starting noise, with a paired 95\% bootstrap interval.
We select each checkpoint on the reward's validation set.
The appendix gives the hyperparameters (\cref{tab:hyperparameters}), full prompts and more samples (\cref{sec:suppl_results}).

\begin{figure*}[t]
\centering
\includegraphics[width=\textwidth]{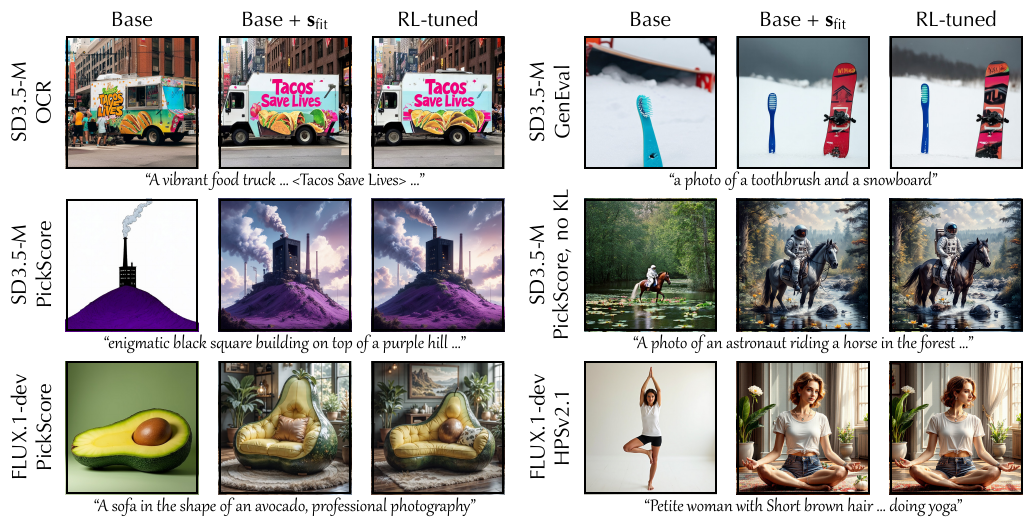}
\caption{\textbf{Steering vectors recover most of the gain of each RL-tuned image model.}
Each prompt's images share one starting noise.}
\label{fig:image-examples}
\end{figure*}
\subsection{Velocity matching works on all eight RL-tuned models}
\label{sec:exp_recovery}
Steering vectors recover 90.0\% to 96.3\% of the reward gain of each of the six RL-tuned image models on the test sets (\cref{tab:recovery-table,fig:image-examples}).

The recovery also holds on prompts that the base model rarely solves.
For each of the four sampling seeds, we take the GenEval test prompts that the base model fails at the other three (146 prompts).
At the held-out seed, the base model succeeds on 14.9\% of them, and the GenEval steering vectors recover 97.7\% of the GenEval-tuned model's $+0.641$ gain.
Likewise, of the test prompt-seed pairs that the base model fails and the RL-tuned model solves, the steering vectors solve 91.4\% on GenEval and 75.3\% on OCR, close to the 93.3\% and 71.6\% that the RL-tuned model itself solves at the next sampling seed (\cref{tab:paired-success}).

\begin{figure*}[t]
\centering
\includegraphics[width=\textwidth]{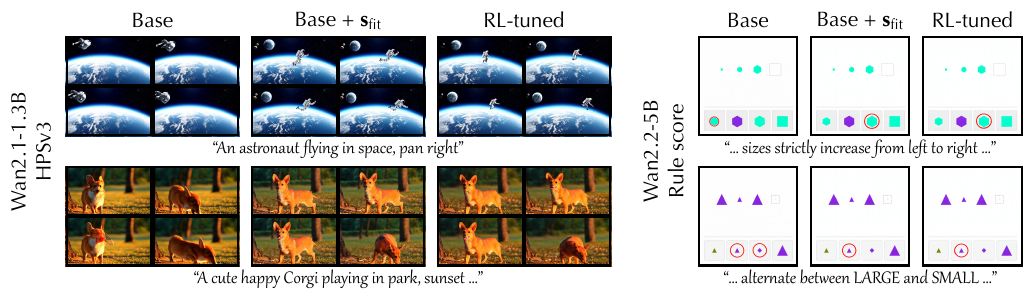}
\caption{\textbf{Steering vectors recover most of the gain of both RL-tuned video models.}
Each prompt's clips share one starting noise.
Wan2.1-1.3B shows four evenly spaced frames per clip, and Wan2.2-5B the last frame, in which the model must circle in red the option that fills the dashed square.}
\label{fig:video-frames}
\end{figure*}
Steering vectors also recover most of the reward gain of the two RL-tuned video models (\cref{tab:recovery-table,fig:video-frames}).
We fit the Wan2.1 steering vectors on 33-frame clips and evaluate them unchanged on 81-frame clips.
On HPSv3, they exceed the RL-tuned model's gain.
The fit matches the RL-tuned model's velocity, not its rewards, so the remaining difference can raise a reward as well as lower it.
The steering vectors also recover most of the gain on HPSv3 percentile and VideoAlign's overall score (\cref{tab:recovery-details}).

\subsection{Fitting finds the change that averaging misses}
\label{sec:ana_distill}
Activation steering usually takes the average change as its constant~\cite{turner2024activation,panickssery2023steering}, and averaging needs no training.
For an RL-tuned model, this average gives the mean-difference steering vectors $\mathbf{s}_{\mathrm{mean}}$ (\cref{sec:method_fit}).
On every model and test set, $\mathbf{s}_{\mathrm{fit}}$ recovers far more of the gain than $\mathbf{s}_{\mathrm{mean}}$ (\cref{tab:recovery-table}).
On PickScore, $\mathbf{s}_{\mathrm{mean}}$ has nearly the right direction but too small a norm: rescaled to the norms of $\mathbf{s}_{\mathrm{fit}}$, it gains $+1.231$, where $\mathbf{s}_{\mathrm{fit}}$ gains $+1.416$.
On OCR, the gain lies off the averaged direction: the component of $\mathbf{s}_{\mathrm{fit}}$ orthogonal to $\mathbf{s}_{\mathrm{mean}}$ keeps 96\% ($+0.312$ of $+0.325$) of the gain of $\mathbf{s}_{\mathrm{fit}}$.

Where the constant enters also matters.
A prompt token fitted to Flow-GRPO models of SD3.5-M recovers 7\% to 32\% of their gains on the PickScore, OCR and GenEval test sets~\cite{chen2026mapreduce}.
Fitted gate scales, a fitted conditioning offset, a fixed prompt suffix and a higher CFG scale gain less than 30\% of what the steering vectors gain (\cref{tab:other-interventions}).

\subsection{Beyond the training reward}
\label{sec:exp_beyond}
\begin{table}[t]
\centering
\caption{\textbf{The steering vectors reproduce effects beyond the training reward.}
The SD3.5-M PickScore-tuned model with KL on the PickScore test set.
Color energy (the summed chroma variance~\cite{pan2026normguard}) and DINOv3~\cite{simeoni2025dinov3} diversity are ratios to the base model.}
\small
\input{tables/beyond-reward}
\label{tab:beyond-reward}
\end{table}

\noindent\textbf{Other scorers.}
The PickScore steering vectors gain about as much as the PickScore-tuned model on two scorers outside the RL objective, HPSv2.1 and ImageReward~\cite{xu2023imagereward} (\cref{tab:beyond-reward}).
Because the PickScore-tuned model also changes image colors strongly (\cref{tab:beyond-reward}), we test whether a global color change accounts for its PickScore gain.
A global color and tone transform, fitted on separate prompts to map the base model's images to that model's colors, recovers only 8\% of that gain on validation prompts.

\noindent\textbf{Diversity and side effects.}
The steered base model's samples are about as diverse as the RL-tuned model's (\cref{tab:beyond-reward}).
The steering vectors of both PickScore-tuned models also reproduce their drop in color energy, and those of the PickScore-tuned model with KL reproduce its narrower range of images for an empty prompt (\cref{tab:beyond-reward,sec:suppl_side_effects}).

\begin{figure}[t]
\centering
\subcaptionbox{Test sets.\label{fig:input-dependence}}{\includegraphics{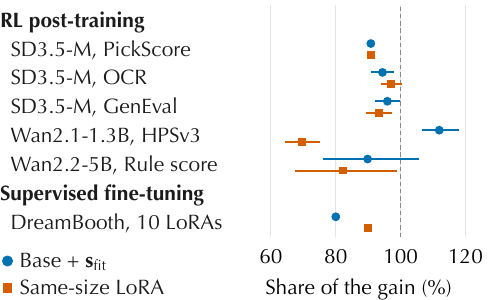}}\hfill
\subcaptionbox{Vectors fitted only to the listed SD3.5-M parts.\label{fig:gain-localisation}}{\includegraphics{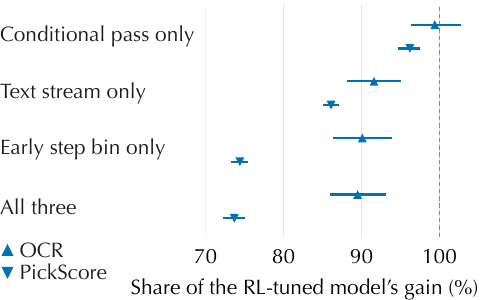}}
\caption{\textbf{(a) A same-size LoRA recovers detectably more after supervised fine-tuning (DreamBooth) but not after RL, and (b) vectors fitted to parts of SD3.5-M recover most of the RL gain.}
Bars are 95\% intervals, and dashed lines mark the fine-tuned model's gain.}
\label{fig:recovery-shares}
\end{figure}

\subsection{Steering vectors vs.\ a same-size LoRA}

\noindent\textbf{A same-size input-dependent fit shows no detected advantage.}
\label{sec:exp_input}
We compare the steering vectors with a same-size LoRA, an input-dependent parameterization with as many parameters as $\mathbf{s}$, fitted by \cref{eq:fit} on the same cached samples for the same number of updates (\cref{fig:input-dependence}).
On the SD3.5-M, VBVR-Pro and FLUX.1-dev models, the paired 95\% interval of the difference between the LoRA and the steering vectors includes zero (\cref{tab:controls-table}).
On Wan2.1, the same-size LoRA recovers 69.7\% of the RL gain, against 111.9\% for the steering vectors.

\noindent\textbf{Supervised fine-tuning.}
In contrast with RL, steering vectors fall short when supervised fine-tuning teaches the base model a subject it cannot draw.
They recover 68\% to 83\% of the identity gain of DreamBooth~\cite{ruiz2023dreambooth} LoRAs for six fictional people and 73\% to 95\% of the subject-similarity gain of four public DreamBooth LoRAs, and a same-size LoRA recovers more (\cref{fig:input-dependence,sec:suppl_sft}).

\subsection{RL on steering vectors alone exceeds the gain}
\label{sec:rl-steering}
\begin{table}[t]
\centering
\caption{\textbf{RL on the steering vectors alone exceeds the gain of both RL-tuned models.}
Flow-GRPO trains only the steering vectors $\mathbf{s}_{\mathrm{RL}}$ of the frozen SD3.5-M, from zero.
Shares are of the RL-tuned model's test-set gain, with 95\% intervals, and the $\mathbf{s}_{\mathrm{fit}}$ rows repeat \cref{tab:recovery-table}.
Preferred: the steered image's win rate in pairwise comparisons (wins plus half the ties) over 50 test prompts and 25 raters (\cref{sec:suppl_user_study}). All win rates against the base model differ from 50\% ($p < 0.05$, sign test); none against the RL-tuned models do.}
\small
\input{tables/rl-steering-study}
\label{tab:rl-steering}
\end{table}
The steering vectors $\mathbf{s}_{\mathrm{fit}}$ reproduce a gain that RL has already found, so we next test whether policy-gradient RL reaches the gain when the steering vectors are all it can train.
We run Flow-GRPO on frozen SD3.5-M, training only the steering vectors $\mathbf{s}_{\mathrm{RL}}$ from zero, so the training sees the reward but not the RL-tuned model (settings in \cref{tab:rl-hyperparameters}).
On the test sets at two sampling seeds, Base + $\mathbf{s}_{\mathrm{RL}}$ exceeds the gains of both the OCR-tuned and PickScore-tuned models (\cref{tab:rl-steering}).
Like the RL-tuned models, it writes the text that the base model gets wrong and adds the suit that the base model leaves out (\cref{fig:rl-steering-examples}).
The part of the RL gain that $\mathbf{s}_{\mathrm{fit}}$ misses comes from the fitting, not the steering-vector form, because $\mathbf{s}_{\mathrm{RL}}$ exceeds both RL-tuned models' gains.
In a user study, the images of Base + $\mathbf{s}_{\mathrm{fit}}$ and Base + $\mathbf{s}_{\mathrm{RL}}$ have a 60\% to 65\% win rate against the base model's and are on par with the RL-tuned models' (50\% to 52\%, \cref{tab:rl-steering}).

%% file: tables/beyond-reward.tex
\begin{tabular}{@{}lr@{.}lr@{.}l@{}}
\toprule
Measure & \multicolumn{2}{c}{RL-tuned} & \multicolumn{2}{c}{Base + $\mathbf{s}_{\mathrm{fit}}$} \\
\midrule
HPSv2.1 gain & +0 & 0407 & +0 & 0393 \\
ImageReward gain & +0 & 388 & +0 & 400 \\
Color energy & 0 & 550 & 0 & 542 \\
Empty-prompt diversity & 0 & 850 & 0 & 847 \\
Within-prompt diversity & 0 & 517 & 0 & 522 \\
\bottomrule
\end{tabular}

%% file: tables/rl-steering-study.tex
\setlength{\tabcolsep}{3pt}
\begin{tabular}{@{}llr@{\,}lcc@{}}
\toprule
 & & \multicolumn{2}{c}{Share of} & \multicolumn{2}{c}{Preferred} \\
\cmidrule(l){5-6}
Reward & Vectors & \multicolumn{2}{c}{RL gain (\%)} & vs Base & vs RL-tuned \\
\midrule
PickScore & $\mathbf{s}_{\mathrm{fit}}$ & 90.8 & [89.8, 91.7] & 64\% & 51\% \\
 & $\mathbf{s}_{\mathrm{RL}}$ & 103.6 & [102.0, 105.1] & 62\% & 52\% \\
\addlinespace
OCR & $\mathbf{s}_{\mathrm{fit}}$ & 94.4 & [90.8, 97.9] & 60\% & 50\% \\
 & $\mathbf{s}_{\mathrm{RL}}$ & 104.6 & [102.1, 107.3] & 65\% & 52\% \\
\bottomrule
\end{tabular}

%% file: sec/5_analysis.tex
\section{Where does the gain act?}
\label{sec:ana_credit}
To see how one shared shift yields prompt-specific gains, we fit restricted sets of steering vectors on SD3.5-M and test how the change crosses joint attention.

\noindent\textbf{Finding 1. The gain favors the conditional pass, the early steps and the text stream.}
We fit each restricted set with the recipe and training length of the full set and evaluate it on the OCR and PickScore test sets at one sampling seed (\cref{fig:gain-localisation}).
Steering vectors that act only on the conditional pass recover more of the RL gains than the full set (\cref{fig:gain-localisation}), whereas the unconditional-pass vectors of a full set, added alone, lower both rewards (\cref{tab:part-ablation-probe}).
The vectors of the two passes point in similar directions and guidance subtracts the unconditional pass, so without its vectors the net guided change grows, as with a larger $\alpha$.
Fitted alone, either stream recovers most of the gain (91.4\% on OCR and 84.9\% on PickScore for the image stream).
Within the full set, however, the text-stream vectors carry most of the gain when added alone (\cref{tab:part-ablation-probe}).
The early step bin carries far more of the gain than the late one.
With all three choices together, 35,328 parameters (12\% of the full set, 0.19\% of the rank-32 LoRA) recover 89.5\% of the RL gain on OCR and 73.7\% on PickScore.

\begin{figure}[t]
\centering
\includegraphics[width=\columnwidth]{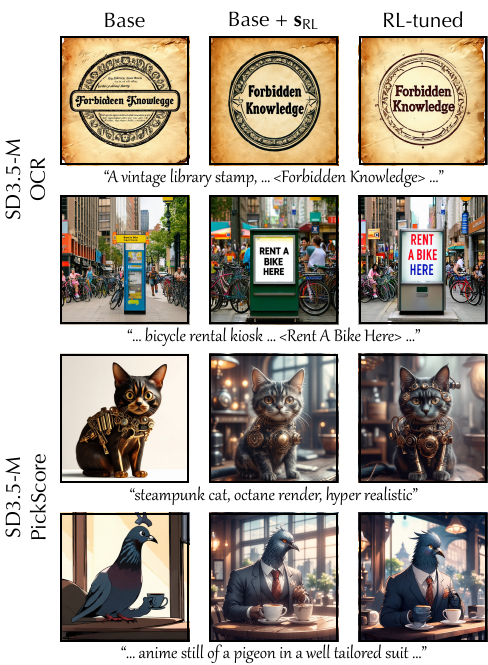}
\caption{\textbf{Steering vectors trained by RL alone change the images as the RL-tuned models do.}
Each prompt's images share one starting noise.}
\label{fig:rl-steering-examples}
\end{figure}
\begin{figure}[t]
\centering
\includegraphics{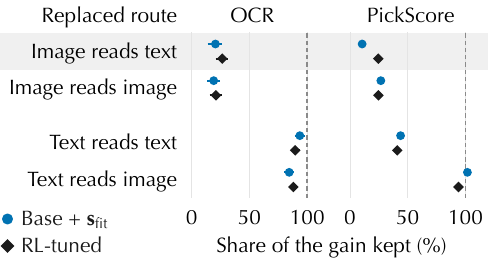}
\caption{\textbf{Steering vectors and RL reach the image through what image tokens read from text.}
Share of the gain kept when one route reads the base model's keys and values, with 95\% intervals.
Image reads image carries the image stream's own changes and what image tokens read from text in earlier blocks.}
\label{fig:attention-route-a}
\end{figure}

\noindent\textbf{Finding 2. The gain reaches the image through what image tokens read from text.}
The text-stream vectors shift text tokens only, and the velocity is projected from the image stream, so their change must reach the image through joint attention~\cite{esser2024scaling}, in which image and text tokens read the keys and values of both streams.
Replacing the text keys and values that image queries read with those of the unsteered base model on the same noisy state keeps only 20.5\% of the steering vectors' OCR gain and 10.5\% of their PickScore gain (\cref{fig:attention-route-a}).
The RL-tuned models rely on the same route, retaining 26.4\% (OCR-tuned) and 24.5\% (PickScore-tuned) of their gains under this intervention, whereas resetting what text tokens read from image tokens preserves at least 84.4\% of all four gains.
On the OCR test prompt of \cref{fig:attention-route-b}, reverting what image tokens read from text turns both models' clean text back into the base model's garbled text.
On PickScore, replacing what text tokens read from other text tokens leaves only 43.8\% of the steering vectors' gain and 40.9\% of the RL-tuned model's, versus 93.6\% and 89.6\% on OCR; preference gains thus also modify text tokens before image tokens read them, consistent with style reaching the image through text tokens~\cite{ge2026vision}.

With text keys and values from the OCR-steered base model, the OCR-tuned model keeps 95.2\% of its gain, against 44.0\% from the PickScore-steered one (\cref{sec:suppl_results}).

\noindent\textbf{How a shared shift becomes prompt-specific.}
The frozen blocks turn the same shift into a different change for each token (\cref{fig:shift-to-velocity}), and one joint-attention head of dimension $d_h$ shows how.
Let image token $i$ with query $\mathbf{q}_i$ put the attention share $m_i$ on the text tokens, and let $\bar{\mathbf{o}}^{\mathrm{txt}}_i$ and $\bar{\mathbf{o}}^{\mathrm{img}}_i$ be its attention-weighted mean values within each stream.
If the query and the image keys and values stay fixed, and every text key moves by one shared vector $\boldsymbol{\delta}_{\mathrm{key}}$ and every text value by $\boldsymbol{\delta}_{\mathrm{val}}$ after normalization, the share and the head output $\mathbf{o}_i$ change to (\cref{thm:attention})
\begin{gather}
  \frac{m_i'}{1 - m_i'} = \frac{m_i}{1 - m_i}\, e^{\mathbf{q}_i^{\top}\boldsymbol{\delta}_{\mathrm{key}}/\sqrt{d_h}}, \nonumber \\
  \mathbf{o}_i' - \mathbf{o}_i = (m_i' - m_i)\big(\bar{\mathbf{o}}^{\mathrm{txt}}_i - \bar{\mathbf{o}}^{\mathrm{img}}_i\big) + m_i'\, \boldsymbol{\delta}_{\mathrm{val}}.
  \label{eq:attention-shift-main}
\end{gather}
Both terms can depend on the prompt via $\mathbf{q}_i$ and $m_i$, while only the first uses the stream averages; without a key shift, $\mathbf{o}_i' - \mathbf{o}_i = m_i\boldsymbol{\delta}_{\mathrm{val}}$.
In SD3.5-M, normalization before attention generally makes the key and value shifts differ across tokens, and \cref{thm:attention} gives their first-order effect.
Cross-attention, in which each query reads text tokens only, differs: a shared key shift after normalization cancels, and a shared value shift adds one fixed vector to every query token.
On Wan2.1, a constant on the key input alone recovers none of the HPSv3 gain, close to this prediction although Wan2.1 normalizes its keys (\cref{sec:suppl_results}).

%% file: sec/6_conclusion.tex
\section{Conclusion}
\label{sec:conclusion}
RL post-training updates the weights of a diffusion transformer, so it changes the activations differently for each prompt and token, as prompt-specific skills such as text rendering seem to require.
Yet steering vectors shared across all inputs and fitted with no reward in the objective recover most of the reward gain of six image and two video RL-tuned models.
Trained with Flow-GRPO on a frozen SD3.5-M, the vectors exceed the reward gains of the OCR-tuned and PickScore-tuned models, and a user study rates the resulting images on par with those of the RL-tuned models.
Most of the RL gain thus lies in a small set of constant vectors, which the frozen generator turns into prompt-specific images.
On SD3.5-M, the gain favors the conditional pass, the early steps and the text stream, and reaches the image through what image tokens read from text in joint attention.
We hypothesize that policy-gradient RL post-training primarily steers the base model toward what it can already compute.
It remains open whether optimizing these rewards benefits from new input-dependent transformations at all, or whether current RL methods cannot learn them.
All eight RL-tuned models use policy gradients, so preference optimization and reward backpropagation remain untested.

%% file: sec/X_suppl.tex
\clearpage
\setcounter{dbltopnumber}{2}
\twocolumn[{\headfont\bfseries\LARGE Appendix\par\vspace{1.0em}}]
\setcounter{section}{0}
\renewcommand{\thesection}{\Alph{section}}
\renewcommand{\theHsection}{suppl.\Alph{section}}
\setcounter{figure}{0}
\renewcommand{\thefigure}{S\arabic{figure}}
\renewcommand{\theHfigure}{suppl.\arabic{figure}}
\setcounter{table}{0}
\renewcommand{\thetable}{S\arabic{table}}
\renewcommand{\theHtable}{suppl.\arabic{table}}
\setcounter{equation}{0}
\renewcommand{\theequation}{S\arabic{equation}}
\renewcommand{\theHequation}{suppl.\arabic{equation}}

Every interval is a paired 95\% bootstrap interval over prompts with 2,000 resamples, or with 10,000 resamples for video, over prompts on Wan2.1 and over instances on VBVR-Pro.

\section{Fitting recipe}
\label{sec:suppl_recipe}

\Cref{tab:hyperparameters} gives all fitting hyperparameters and the fitting cost.
Every fit uses AdamW with betas (0.9, 0.999), eps $10^{-8}$, no weight decay and gradient clipping at norm 1, with one training seed.
It starts at $\mathbf{s}_{\mathrm{mean}}$, measured on 256 calibration prompts.
The training prompts are those of Flow-GRPO~\cite{liu2026flow} on SD3.5-M and FLUX.1-dev, of GenRL~\cite{genrl} on Wan2.1, and of the public VBVR-Pro RL training set~\cite{xu2026vbvr} on Wan2.2-5B.
On SD3.5-M, the calibration and validation prompts come from the same training files.
The learning rate rises linearly over the first 50 updates and stays constant until its decay.
For the OCR-tuned and GenEval-tuned models, the checkpoint at 3,000 updates comes before the decay.
New training states re-noise final images at a level drawn uniformly from the sampler's 40 noise levels, with new noise at each update.
Fixed training states re-noise each prompt's final image once per contiguous range of levels, with fixed noise.
\begin{table*}[t]
\centering
\caption{\textbf{Fitting hyperparameters and cost of the steering vectors for the RL-tuned models of \cref{tab:recovery-table}.}
The learning rate decays linearly to 10\% of its peak from the listed update to the last.
GPU-hours count sampling the RL-tuned model, measuring $\mathbf{s}_{\mathrm{mean}}$ and training.}
\small
\setlength{\tabcolsep}{5pt}
\input{tables/hyperparameters}
\label{tab:hyperparameters}
\end{table*}

\Cref{tab:rl-hyperparameters} gives the training settings of the steering vectors that Flow-GRPO trains from zero (\cref{sec:rl-steering}).
\begin{table}[t]
\centering
\caption{\textbf{Training settings of Flow-GRPO on the steering vectors $\mathbf{s}_{\mathrm{RL}}$ of \cref{tab:rl-steering}.}
Other settings follow the PickScore and OCR recipes of Flow-GRPO~\cite{liu2026flow}.}
\small
\begin{tabular}{@{}lcc@{}}
\toprule
 & PickScore & OCR \\
\midrule
Learning rate & $5{\times}10^{-3}$& $5{\times}10^{-3}$\\
Images per update &576&576\\
Training updates &1{,}300&1{,}500\\
B200 GPU-hours & 123&183 \\
EMA & \multicolumn{2}{c}{decay 0.9, updated every 8 updates} \\
\bottomrule
\end{tabular}
\label{tab:rl-hyperparameters}
\end{table}

\paragraph{Training length.}
On SD3.5-M, we choose the number of training updates for each RL-tuned model among the checkpoints at 1,500, 3,000 and 4,500 updates of one training run, by the reward gain on that RL-tuned model's validation set.
No choice uses the test sets.
The fitting loss on held-out prompts, which uses no reward, is lowest at 4,500 updates for all four SD3.5-M RL-tuned models.
Choosing the length by this loss keeps the length of the two PickScore-tuned models, and it changes the validation gain of the OCR-tuned and GenEval-tuned models by $-0.011$ [$-0.026$, $+0.004$] and $-0.021$ [$-0.059$, $+0.017$].

\paragraph{Steering-vector layouts.}
\Cref{tab:shift-layouts} gives the layout of the steering vectors on every base model.
\begin{table}[t]
\centering
\caption{\textbf{The steering vectors on every base model.}
Parameters are the block outputs times the width, the passes and two step bins.
FLUX.1-dev is guidance-distilled and the VBVR-Pro fine-tune of Wan2.2-5B samples without guidance, so both run one pass.}
\small
\input{tables/shift-layouts}
\label{tab:shift-layouts}
\end{table}

\paragraph{Step bins.}
The energy of an RL-tuned model's average change to the block outputs is its squared norm summed over the sampling steps.
The change is in units of the base model's activation standard deviation on SD3.5-M and in raw units on FLUX.1-dev.
Averaging the change within contiguous groups of sampling steps keeps part of this energy.
\Cref{tab:step-bins} gives, for every image RL-tuned model, the energy kept with two step bins as a share of the energy kept with eight contiguous groups.
On SD3.5-M with the PickScore-tuned model, the change has cosine similarity 0.09 between the first five and the last five sampling steps.
\begin{table}[t]
\centering
\caption{\textbf{Two step bins keep most of the energy of each RL-tuned model's change over the sampling steps.}
Kept is relative to eight contiguous groups of sampling steps.
For unrelated steps it is 25\% on SD3.5-M and 25\% to 26\% on FLUX.1-dev.}
\small
\input{tables/step-bins}
\label{tab:step-bins}
\end{table}

\section{Properties of the fit}
\label{sec:suppl_fit}

\begin{figure*}[t]
\centering
\includegraphics{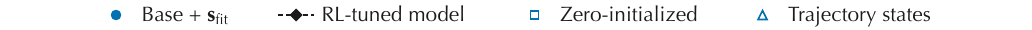}\\[2pt]
\subcaptionbox{\label{fig:fit-sensitivity-a}}{\includegraphics{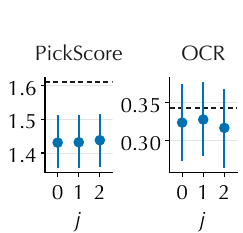}}\hfill
\subcaptionbox{\label{fig:fit-sensitivity-b}}{\includegraphics{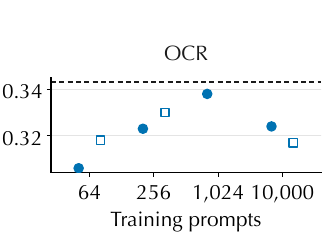}}\hfill
\subcaptionbox{\label{fig:fit-sensitivity-c}}{\includegraphics{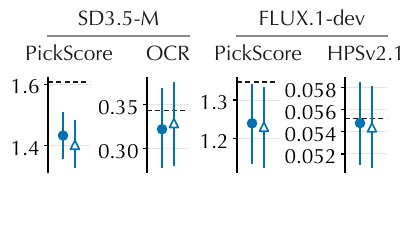}}\\[2pt]
\subcaptionbox{\label{fig:fit-sensitivity-d}}{\includegraphics{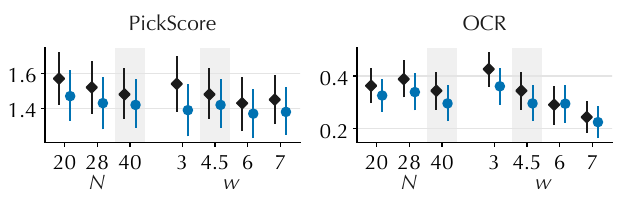}}\hfill
\subcaptionbox{\label{fig:fit-sensitivity-e}}{\includegraphics{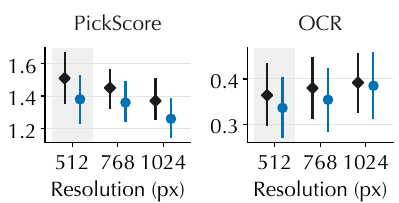}}
\caption{\textbf{The steering vectors recover most of their RL-tuned model's gain across training seeds, prompt counts, training states, sampler settings and resolutions.}
Raw gain over the base model with 95\% intervals, dashed lines for the RL-tuned model's gain and grey bands for the training setting.
(a)~Training seed $j$.
(b)~Training prompts.
(c)~Training states.
(d)~Sampling steps $N$ and guidance scale $w$.
(e)~Resolution.}
\label{fig:fit-sensitivity}
\end{figure*}

\paragraph{Training seeds.}
For each of the PickScore-tuned and OCR-tuned models on SD3.5-M, fits with three training seeds reach nearly the same gain.
After 1,500 training updates, the three PickScore gains span 0.006 points on 512 PickScore prompts (\cref{fig:fit-sensitivity}).
The three OCR gains span 0.011 on 256 OCR prompts.

\paragraph{Starting point.}
Zero initialization reaches nearly the same gain as initialization at $\mathbf{s}_{\mathrm{mean}}$ on PickScore and OCR.
On the PickScore test subset of 512 prompts, a zero-initialized fit gains $+1.34$ against $+1.36$ for a fit initialized at $\mathbf{s}_{\mathrm{mean}}$.
On 256 OCR prompts, the zero-initialized fit on 10,000 training prompts gains $+0.317$ and the fit initialized at $\mathbf{s}_{\mathrm{mean}}$ gains $+0.324$.
The two OCR fits reach this gain with different vectors, which have cosine similarity 0.33.
The fit initialized at $\mathbf{s}_{\mathrm{mean}}$ keeps cosine similarity 0.97 with $\mathbf{s}_{\mathrm{mean}}$, and the change that training adds to it has cosine similarity 0.90 with the zero-initialized fit.
All four fits use at most 1,500 training updates and one training seed.

\paragraph{Prompt count.}
On OCR, 64 distinct training prompts suffice to fit the steering vectors (\cref{fig:fit-sensitivity}).
Every fit in this ablation uses about 10,000 samples of the RL-tuned model, at most 1,500 training updates and one training seed.
On 256 OCR prompts, a fit initialized at $\mathbf{s}_{\mathrm{mean}}$ on 256 training prompts keeps the full gain of a fit on 10,000 prompts.
A fit on 64 prompts keeps 94\% of that gain, and a zero-initialized fit on 64 prompts keeps all of its own 10,000-prompt gain.
The zero-initialized fit on 64 prompts gains $+0.318$, and its paired difference to the zero-initialized fit on 10,000 prompts is $+0.000$ [$-0.019$, $+0.019$].
At every prompt count, each fit's gain lies within 0.02 of its own 10,000-prompt gain.

\begin{figure*}[t]
\centering
\includegraphics[width=\textwidth]{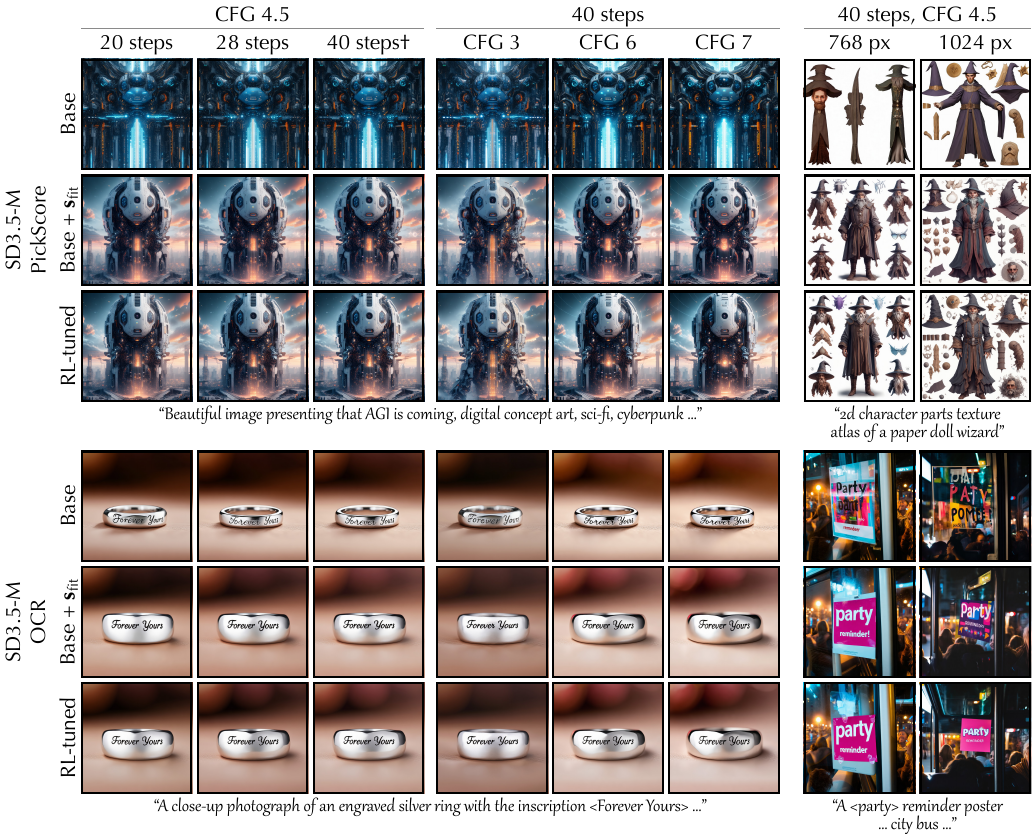}
\caption{\textbf{Steering vectors fitted at one sampler setting transfer to other settings and resolutions.}
Each column shares one prompt and starting noise.
$\dagger$ marks the setting of the fit, which also uses 512\,px.}
\label{fig:transfer-examples}
\end{figure*}
\paragraph{Training states.}
The steered base model should match the RL-tuned model on the states that the RL-tuned model's own sampling visits.
Velocity matching uses re-noised final images of the RL-tuned model, which lie near its sampling trajectory.
For each of four RL-tuned models on SD3.5-M and FLUX.1-dev, we fit one set of steering vectors on its re-noised final images and one on states from its sampling trajectory.
Each fit uses 1,500 training updates and one training seed, and each is scored on its reward's prompts with one sampling seed.
On all four, the two fits reach nearly the same gain (\cref{fig:fit-sensitivity}).
The largest difference, on SD3.5-M with the PickScore-tuned model, is $-0.030$ [$-0.050$, $-0.011$] points on 512 PickScore prompts, sampling-trajectory states minus re-noised states.

\paragraph{Sampler settings and resolution.}
We fit the PickScore and OCR steering vectors at 40 sampling steps, CFG 4.5 and 512\,px.
At every tested setting, we compare each set of steering vectors with its RL-tuned model sampled at that setting.
The PickScore steering vectors use 4,500 training updates and the OCR steering vectors 1,500, each with one training seed.
The sampler-setting comparison uses 128 test prompts per reward.
Across 20 and 28 sampling steps and CFG 3, 6 and 7, the PickScore steering vectors recover 90\% to 96\% of their RL-tuned model's gain (\cref{fig:fit-sensitivity}).
The OCR steering vectors recover 85\% to 101\% of their RL-tuned model's gain.
The resolution comparison uses 128 prompts per reward.
At 768 and 1024\,px, the PickScore steering vectors recover 92\% to 94\% of their RL-tuned model's gain and the OCR steering vectors 93\% to 98\% (\cref{fig:fit-sensitivity}).
\Cref{fig:transfer-examples} shows examples.

\section{Side effects}
\label{sec:suppl_side_effects}
\begin{figure}[t]
\centering
\includegraphics{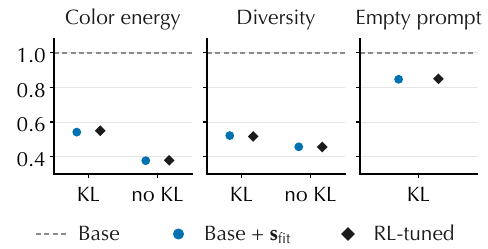}
\caption{\textbf{The steering vectors reproduce the loss of color energy and diversity of both PickScore-tuned models.}
SD3.5-M, as ratios to the base model on the PickScore test set.
Bars, mostly hidden by the markers, are 95\% intervals where measured.}
\label{fig:side-effects}
\end{figure}
\begin{figure}[t]
\centering
\includegraphics[width=\linewidth]{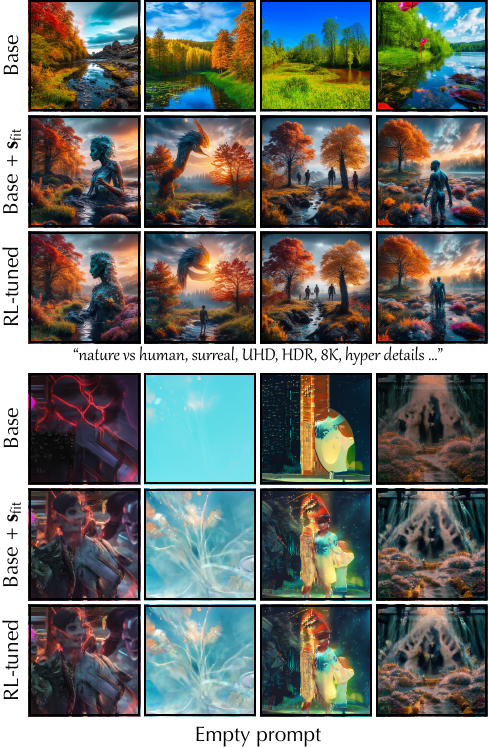}
\caption{\textbf{On PickScore, the steered base model draws nearly the same images as the RL-tuned model from the same noise.}
SD3.5-M with the PickScore-tuned model, on one prompt (top) and the empty prompt (bottom).
Per block, each column shares a seed.}
\label{fig:side-effect-examples}
\end{figure}
On SD3.5-M, the steering vectors reproduce the side effects of both Flow-GRPO PickScore-tuned models, trained with and without the KL penalty (\cref{fig:side-effects}).
We fit both sets of steering vectors for 4,500 training updates, the length selected on the PickScore validation set, with one training seed.
We report color energy as a ratio to the base model's value on the PickScore test set of 2,048 prompts.
\Cref{fig:side-effect-examples} shows examples.

\paragraph{Color.}
Color energy is the summed variance of an image's two chroma channels in YCbCr~\cite{pan2026normguard}.
Both RL-tuned models lower the color energy of the base model's images, and the RL-tuned model without KL lowers it more than the RL-tuned model with KL.
Each set of steering vectors follows its own RL-tuned model.
The RL-tuned model with KL keeps 0.550 of the base model's color energy, and the steered base model 0.542.
The RL-tuned model without KL keeps 0.379, and the steered base model 0.377.

\paragraph{Diversity and the empty prompt.}
Both RL-tuned models also narrow the range of images that the base model draws for one prompt, and the steering vectors narrow it by the same amount (\cref{fig:side-effects}).
Across two sampling seeds on the PickScore test set of 2,048 prompts, the RL-tuned model with KL keeps 0.517 of the base model's within-prompt DINOv3~\cite{simeoni2025dinov3} diversity, and the steered base model 0.522.
Within-prompt diversity is the mean cosine distance between the DINOv3 class-token embeddings of a prompt's images at different sampling seeds.
Across three sampling seeds on the same prompts, the RL-tuned model without KL keeps 0.456 [0.445, 0.466], and the steered base model 0.457 [0.447, 0.467], while recovering 95.5\% [94.9, 96.1] of its gain.
The RL-tuned model with KL also narrows the images it draws for the empty prompt.
Over 128 sampling seeds of the empty prompt, with a bootstrap over the seeds, it keeps 0.850 [0.826, 0.874] of the base model's DINOv3 diversity, and the steered base model 0.847 [0.823, 0.871].

\section{Controls}
\label{sec:suppl_controls}
\begin{table*}[t]
\centering
\caption{\textbf{A same-size LoRA shows no detected advantage over the steering vectors.}
The same-size LoRA sits on attention output projections, with rank 2 on the image models and rank 1 on the video models, so it has as many parameters as $\mathbf{s}$: 288,768 on SD3.5-M, 466,944 on FLUX.1-dev and 184,320 on Wan2.1 and Wan2.2-5B.
The last column is in raw reward points.}
\small
\setlength{\tabcolsep}{4.5pt}
\input{tables/controls-table}
\label{tab:controls-table}
\end{table*}
\Cref{tab:controls-table} compares the steering vectors with a same-size LoRA, fitted by \cref{eq:fit} on the same cached samples for the same number of updates.
The SD3.5-M, Wan2.1 and Wan2.2-5B rows use the test sets of \cref{tab:recovery-table}.
A rank-$r$ LoRA adds $r$ fixed directions to each token, each weighted by a linear function of that token's input, so the comparison tests whether input-dependent weights help at this size.
A rank-1 LoRA is thus a steering vector whose weight depends on the input.
On the five image rows and on the in-domain instances of VBVR-Pro, the paired 95\% interval of the difference between the steering vectors and the LoRA includes zero.
On Wan2.1 it recovers 69.7\% of the RL gain, against 111.9\% for the steering vectors.
Truncating the RL-tuned models' own LoRAs tests whether a rank-1 part of the RL weight update carries the gain.
On the test sets at two sampling seeds, the rank-1 LoRAs of the PickScore-tuned and OCR-tuned models gain $+0.668$ and $+0.210$, against $+1.549$ and $+0.334$ for the full LoRAs, so rank 1 keeps only 43.1\% and 62.8\% of the gain.
A low-rank copy of the RL weight update therefore recovers less than the steering vectors.
Averaging the RL-tuned models' own LoRA change tests how much of it is shared across inputs.
At each block, this change is the block output with the LoRA minus the block output without it, both on the RL-tuned model's block input.
We average it over the tokens of 1,024 noisy states from training prompts, in the layout of $\mathbf{s}$, and add only this average to the base model's block outputs.
On two noisy states of each of 512 prompts, the average closes 62.5\% [61.1, 63.8] of the mean squared velocity gap between the base and PickScore-tuned models and 30.8\% [30.0, 31.8] of the gap to the OCR-tuned model, with 95\% bootstrap intervals over prompts.
Adding the full change at every block closes all of it.
The direction of $\mathbf{s}_{\mathrm{fit}}$ also matters: random vectors, each rescaled to the norm of the steering vector it replaces, lower the base model's PickScore by 4.704 points on the PickScore test set, where the steering vectors raise it by 1.416.
A random change of the RL update's size tests whether the fit recovers any small change of the weights as well as it recovers RL's.
In each LoRA layer of the PickScore-tuned and OCR-tuned models, we replace the trained output factor by a Gaussian matrix with 4.5 times its Frobenius norm, which brings the base model's loss in \cref{eq:fit} to 0.957 and 1.586 times its loss to the RL-tuned model.
We fit steering vectors to each random copy with the recipe of \cref{tab:hyperparameters} for 4,500 updates, at one training seed.
On held-out states of the fit, the steering vectors lower the loss of \cref{eq:fit} by 86.5\% for the PickScore-tuned model and 68.8\% for the OCR-tuned model, against 44.0\% and 32.6\% for the random copies.
On separately drawn noisy states from other prompts, the steering vectors close 88.4\% [87.2, 89.4] of the mean squared velocity gap between the base and PickScore-tuned models, and 68.9\% [67.0, 70.7] of the gap to the OCR-tuned model, so they recover the reward gain without matching the RL-tuned model's velocity exactly.
The fit thus matches the change that RL makes about twice as closely as a random change of the same layout and size.

\begin{table}[t]
\centering
\caption{\textbf{Other input-independent interventions gain less than the steering vectors.}
The gate scales are the gate scalars of Calibri~\cite{tokhchukov2026calibri}, and the conditioning offset is added to the conditioning embedding of the adaptive normalization layers.
The prompt suffix appends quality words or ``with clear, legible text''.}
\small
\input{tables/other-interventions}
\label{tab:other-interventions}
\end{table}
\Cref{tab:other-interventions} compares the steering vectors with other interventions that are the same for every prompt.
The gate scales, the conditioning offset and the steering vectors of this table are fitted by \cref{eq:fit} for 1,500 updates and evaluated on SD3.5-M at one sampling seed.
The CFG rows sample the base model at that scale instead of 4.5.

\section{Supervised fine-tuning}
\label{sec:suppl_sft}
\begin{figure*}[t]
\centering
\includegraphics{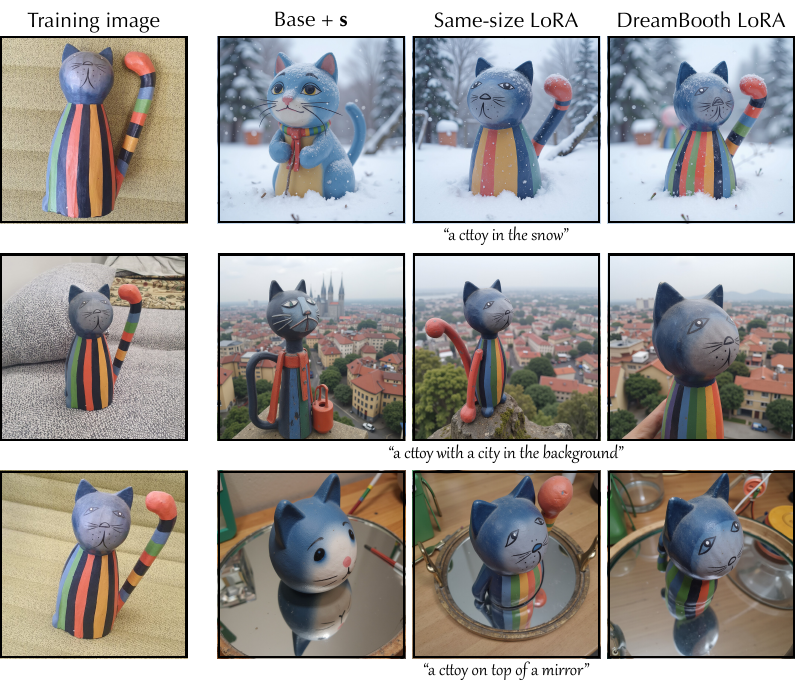}
\caption{\textbf{The steering vectors draw a new subject's coarse appearance and miss its details, which the same-size LoRA keeps.}
FLUX.1-dev models on held-out prompts with the trigger word cttoy, with one starting noise per row.
Base + $\mathbf{s}$ and the same-size LoRA are fitted to the DreamBooth LoRA.}
\label{fig:dreambooth-examples}
\end{figure*}
\begin{figure*}[t]
\centering
\includegraphics{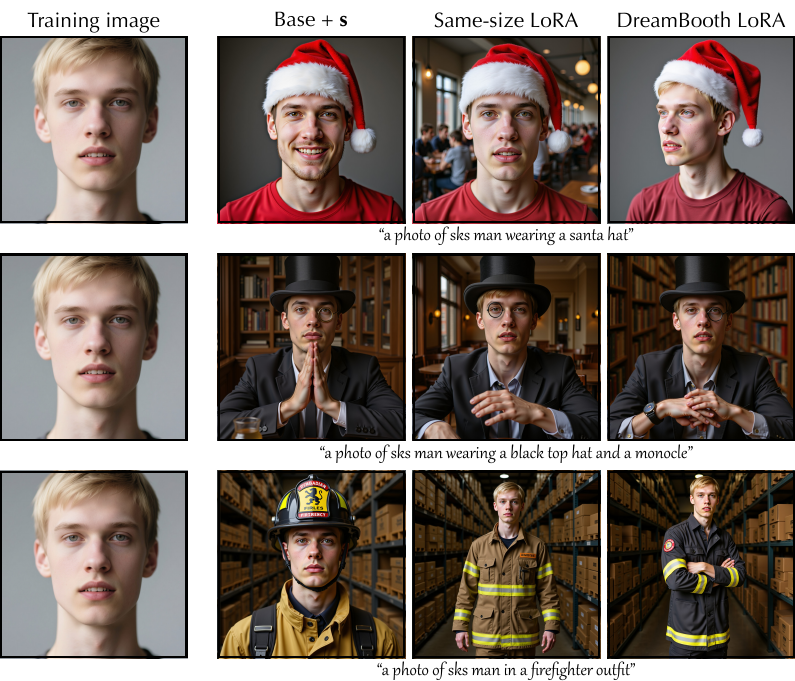}
\caption{\textbf{The steering vectors can draw a fictional person's coarse look but another identity, which the same-size LoRA keeps.}
FLUX.1-dev models of one of the six people on held-out prompts with the trigger word sks, one starting noise per row.
Base + $\mathbf{s}$ and the same-size LoRA are fitted to the person's DreamBooth LoRA.
\cref{tab:faces} gives all six people.}
\label{fig:faces-examples}
\end{figure*}
\begin{table}[t]
\centering
\caption{\textbf{Steering vectors recover part of the identity gain of DreamBooth on six fictional people.}
Share of the DreamBooth LoRA's ArcFace similarity gain over FLUX.1-dev in \%, with 95\% intervals over the 25 held-out prompts.}
\label{tab:faces}
\small
\begin{tabular}{lcc}
\toprule
Person & Base + $\mathbf{s}$ & Same-size LoRA \\
\midrule
1 & 68 [65, 70] & 82 [80, 84] \\
2 & 71 [68, 74] & 85 [83, 86] \\
3 & 73 [70, 75] & 81 [79, 83] \\
4 & 79 [77, 81] & 89 [85, 92] \\
5 & 83 [82, 85] & 91 [90, 93] \\
6 & 71 [68, 74] & 85 [82, 87] \\
\bottomrule
\end{tabular}
\end{table}
DreamBooth~\cite{ruiz2023dreambooth} teaches a model a new subject by supervised fine-tuning on a few images, not by RL.
We fit steering vectors to four public DreamBooth LoRAs whose subjects the base model cannot draw, one on SD3.5-M and three on FLUX.1-dev.
Each fit follows the main recipe for 1,500 updates with one training seed.
Subject similarity is the DINOv3~\cite{simeoni2025dinov3} similarity of an image to the closest training image of the subject.
On 25 held-out prompts with the trigger word at four sampling seeds, the steering vectors reach 73\% to 95\% of the DreamBooth LoRA's similarity gain over the base model.
They reproduce the subject's coarse appearance but miss fine identity details, such as the painted face of a cat toy (\cref{fig:dreambooth-examples}) and the thorns of a cactus character.
A LoRA with as many parameters as the steering vectors reaches 94\% to 100\% of the gain and keeps these details more often.
Without the trigger word, the steering vectors' similarity gain is below the DreamBooth LoRA's own or not detectably different from it.
A public subject may already be partly known to the base model.
We therefore also create six fictional people, three women and three men.
Qwen-Image~\cite{wu2025qwen} draws one portrait of each person from a written description, and FLUX.1 Kontext~\cite{labs2025flux} redraws that person in nine further scenes.
A DreamBooth LoRA of FLUX.1-dev learns each person from these ten images, and the steering vectors and the same-size LoRA are fitted to it as above.
Identity similarity is the ArcFace~\cite{deng2019arcface} cosine similarity of the largest face in an image to the closest training image.
On the 25 held-out prompts at four sampling seeds, the steering vectors recover 68\% to 83\% of the DreamBooth LoRA's identity gain, and the same-size LoRA 81\% to 91\% (\cref{tab:faces,fig:faces-examples}).
Supervised fine-tuning on a new subject thus needs more than a constant shift, whereas RL post-training does not (\cref{sec:exp_recovery}).

\section{The attention identity}
\label{sec:suppl_proofs}
\setcounter{theorem}{0}
\renewcommand{\thetheorem}{S\arabic{theorem}}
\providecommand{\theHtheorem}{}
\renewcommand{\theHtheorem}{suppl.\arabic{theorem}}
This section proves the attention identity of \cref{sec:ana_credit}.

\begin{figure}[t]
\centering
\includegraphics{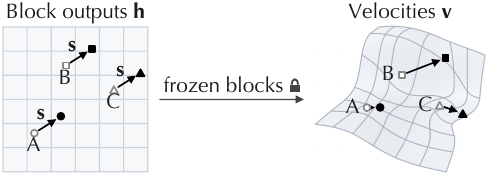}
\caption{\textbf{The frozen blocks turn the same shift into different velocity changes.}
A sketch: the same $\mathbf{s}$ moves the block outputs A, B and C equally, and their velocities change by different amounts and in different directions.}
\label{fig:shift-to-velocity}
\end{figure}

\paragraph{Joint attention with shifted keys and values.}
Consider one head of joint attention with head dimension $d_h$, as in \cref{sec:ana_credit}.
Image token $i$ has the query $\mathbf{q}_i$, and token $j$ of either stream has the key $\mathbf{k}_j$ and the value $\mathbf{u}_j$, which we do not call $\mathbf{v}_j$, since $\mathbf{v}$ is the velocity.
These are the vectors that enter the attention, after any normalization.
The attention probabilities are
\begin{equation}
  m_{ij} = \frac{e^{\mathbf{q}_i^{\top}\mathbf{k}_j/\sqrt{d_h}}}{\sum_{j'} e^{\mathbf{q}_i^{\top}\mathbf{k}_{j'}/\sqrt{d_h}}} ,
  \label{eq:attention-weights}
\end{equation}
where $j'$ runs over the tokens of both streams, and the head outputs $\mathbf{o}_i = \sum_j m_{ij}\, \mathbf{u}_j$.
Let $m_i = \sum_{j \in \mathrm{txt}} m_{ij}$ be the attention share on the text tokens, which lies in $(0, 1)$ in joint attention.
Let $\bar{\mathbf{o}}^{\mathrm{txt}}_i = \sum_{j \in \mathrm{txt}} m_{ij}\, \mathbf{u}_j/m_i$ and $\bar{\mathbf{o}}^{\mathrm{img}}_i = \sum_{j \in \mathrm{img}} m_{ij}\, \mathbf{u}_j/(1 - m_i)$ be the averages of the values within each stream, so that $\mathbf{o}_i = m_i\, \bar{\mathbf{o}}^{\mathrm{txt}}_i + (1 - m_i)\, \bar{\mathbf{o}}^{\mathrm{img}}_i$.
Primes mark the quantities after a change.
If every text key moves by one vector $\boldsymbol{\delta}_{\mathrm{key}}$ and every text value by one vector $\boldsymbol{\delta}_{\mathrm{val}}$, with the query and the image tokens unchanged, then
\begin{gather}
  \frac{m_i'}{1 - m_i'} = \frac{m_i}{1 - m_i}\, e^{\mathbf{q}_i^{\top}\boldsymbol{\delta}_{\mathrm{key}}/\sqrt{d_h}}, \nonumber \\
  \mathbf{o}_i' - \mathbf{o}_i = (m_i' - m_i)\big(\bar{\mathbf{o}}^{\mathrm{txt}}_i - \bar{\mathbf{o}}^{\mathrm{img}}_i\big) + m_i'\, \boldsymbol{\delta}_{\mathrm{val}}.
  \label{eq:attention-shift}
\end{gather}
\begin{theorem}[Attention with shifted keys and values]
\label{thm:attention}
(i) Let every text key change to $\mathbf{k}_j + \boldsymbol{\delta}_{\mathrm{key}}$ and every text value to $\mathbf{u}_j + \boldsymbol{\delta}_{\mathrm{val}}$, with the query, the image keys and the image values unchanged.
Then the weights $m_{ij}/m_i$ within the text tokens and $m_{ij}/(1 - m_i)$ within the image tokens are unchanged, and \cref{eq:attention-shift} holds.
(ii) For changes $d\mathbf{q}_i$, $d\mathbf{k}_j$ and $d\mathbf{u}_j$ that may differ across tokens, the output changes to first order by
\begin{multline}
  d\mathbf{o}_i = \sum_j m_{ij}\, d\mathbf{u}_j \\
  + \frac{1}{\sqrt{d_h}} \sum_j m_{ij}\, (\mathbf{u}_j - \mathbf{o}_i)\, \big(\mathbf{q}_i^{\top} d\mathbf{k}_j + \mathbf{k}_j^{\top} d\mathbf{q}_i\big) .
  \label{eq:attention-differential}
\end{multline}
(iii) If image token $i$ reads the text tokens only, as in cross-attention, and its query is unchanged, a common shift of the text keys leaves its output unchanged, and a common shift of the text values adds $\boldsymbol{\delta}_{\mathrm{val}}$ to it.
\end{theorem}
\emph{Proof.}
(i) Let $S^{\mathrm{txt}}_i$ and $S^{\mathrm{img}}_i$ be the sums of the numerators of \cref{eq:attention-weights} over the text and the image tokens, so that $m_i/(1 - m_i) = S^{\mathrm{txt}}_i/S^{\mathrm{img}}_i$.
The key shift multiplies the numerator of every text token by $e^{\mathbf{q}_i^{\top}\boldsymbol{\delta}_{\mathrm{key}}/\sqrt{d_h}}$ and leaves those of the image tokens unchanged.
The weights within one stream are ratios of numerators of that stream, so they are unchanged.
$S^{\mathrm{txt}}_i$ is multiplied by the same factor and $S^{\mathrm{img}}_i$ is unchanged, which gives the first line of \cref{eq:attention-shift}.
The weights within the text tokens sum to one, so the value shift moves $\bar{\mathbf{o}}^{\mathrm{txt}}_i$ to $\bar{\mathbf{o}}^{\mathrm{txt}}_i + \boldsymbol{\delta}_{\mathrm{val}}$, and $\bar{\mathbf{o}}^{\mathrm{img}}_i$ is unchanged.
Hence $\mathbf{o}_i' = m_i'\, (\bar{\mathbf{o}}^{\mathrm{txt}}_i + \boldsymbol{\delta}_{\mathrm{val}}) + (1 - m_i')\, \bar{\mathbf{o}}^{\mathrm{img}}_i$, and subtracting $\mathbf{o}_i$ gives the second line.
(ii) The logit $\mathbf{q}_i^{\top}\mathbf{k}_j/\sqrt{d_h}$ changes to first order by $\Delta_{ij} = \big(\mathbf{q}_i^{\top} d\mathbf{k}_j + \mathbf{k}_j^{\top} d\mathbf{q}_i\big)/\sqrt{d_h}$.
Differentiating \cref{eq:attention-weights} gives $dm_{ij} = m_{ij}\big(\Delta_{ij} - \sum_{j'} m_{ij'}\, \Delta_{ij'}\big)$.
Then $d\mathbf{o}_i = \sum_j m_{ij}\, d\mathbf{u}_j + \sum_j dm_{ij}\, \mathbf{u}_j$, and $\sum_j m_{ij}\, \mathbf{u}_j = \mathbf{o}_i$, so $\sum_j dm_{ij}\, \mathbf{u}_j = \sum_j m_{ij}\, \Delta_{ij}\, (\mathbf{u}_j - \mathbf{o}_i)$, which is \cref{eq:attention-differential}.
(iii) If token $i$ reads only text tokens, the key shift multiplies every numerator of \cref{eq:attention-weights} by the same factor, so no weight changes.
The weights sum to one, so the value shift adds $\boldsymbol{\delta}_{\mathrm{val}}$ to the output.
\hfill$\square$

With $d\mathbf{q}_i = \mathbf{0}$, the changes $\boldsymbol{\delta}_{\mathrm{key}}$ and $\boldsymbol{\delta}_{\mathrm{val}}$ for the text tokens and no change for the image tokens, \cref{eq:attention-differential} is the first-order form of \cref{eq:attention-shift}, since $\bar{\mathbf{o}}^{\mathrm{txt}}_i - \mathbf{o}_i = (1 - m_i)\, (\bar{\mathbf{o}}^{\mathrm{txt}}_i - \bar{\mathbf{o}}^{\mathrm{img}}_i)$.
In SD3.5-M, a steering vector added to a text-stream block output reaches the keys and values of the next block through its normalization layers and projections, which generally make $d\mathbf{k}_j$ and $d\mathbf{u}_j$ differ across tokens.
Part (ii) gives the change in this case, and each of its terms can depend on the prompt through $m_{ij}$, $\mathbf{q}_i$, $\mathbf{k}_j$ and $\mathbf{u}_j$.
Petrov \etal~\cite{petrov2024prompting} derive the analogous decomposition for a prefix added to a language model, which scales down the attention to the other tokens without changing its proportions and biases the output.

\section{User study}
\label{sec:suppl_user_study}
\Cref{tab:user-study} splits the overall preferences of \cref{tab:rl-steering} into prompt following and aesthetics. All 25 raters consented to take part and gave anonymous pairwise judgments of generated images. No review board reviewed the study because it collected only these anonymous judgments and kept no personal data.

\begin{table}[h]
\centering
\caption{\textbf{Pairwise user study on SD3.5-M.}
Win rate (\%) of the steered image, wins plus half the ties, for prompt following (PF), aesthetics (Aes) and overall preference (All). Each cell holds about 500 judgments over 50 test prompts and 25 raters. All win rates against the base model differ from 50\% ($p < 0.05$, sign test), and none against the RL-tuned models does.}
\small
\input{tables/user-study}
\label{tab:user-study}
\end{table}

\section{Limitations}
\label{sec:suppl_limitations}
Steering vectors suffice to recover most of each RL-tuned model's reward gain (\cref{sec:exp_recovery}).
We make no claim of necessity, that the RL-tuned model itself needs the steering vectors to reach its gain.
Nor do we claim that RL post-training is low-rank.
The steering vectors recover most of the gain of RL-tuned models trained as LoRAs and of two full fine-tunes alike, so their few parameters do not imply a low-rank RL update.
On video, the claim covers two RL-tuned models on two base models, Wan2.1-T2V-1.3B and Wan2.2-TI2V-5B, and not video RL post-training in general.

\section{Additional results}
\label{sec:suppl_results}
\begin{table*}[t]
\centering
\caption{\textbf{On every row and score, the steering vectors recover a larger share of the RL gain than $\mathbf{s}_{\mathrm{mean}}$.}
Gains are raw reward points on the test sets of \cref{tab:recovery-table} and on the VBVR-Pro out-of-domain split.
MQ, TA and VQ are VideoAlign's motion quality, text alignment and visual quality, and overall is their sum.}
\small
\input{tables/recovery-details}
\par\vspace{2pt}{\footnotesize $^{*}$The RL-tuned model did not train on VQ.\par}
\label{tab:recovery-details}
\end{table*}
\begin{table*}[t]
\centering
\caption{\textbf{The eight RL-tuned models span four base models, four RL methods, and LoRA and full fine-tuning.}
Full is a fine-tune of all weights, and FLUX.1-dev takes the guidance scale as an input to one pass.
Wan2.1 samples with UniPC, and VBVR-Pro with Flow-CPS from a given first frame.}
\small
\input{tables/models-and-samplers}
\label{tab:models-and-samplers}
\end{table*}
\begin{table}[t]
\centering
\caption{\textbf{Separate vectors for the two guidance passes raise the gain more than step bins do.}
PickScore gain on SD3.5-M.
Passes is 2 with separate vectors per guidance pass, Streams is the image stream or both streams, and Bins counts the step bins.
The row 2, Both, 2 is the layout of \cref{tab:recovery-table}.}
\small
\setlength{\tabcolsep}{3.5pt}
\input{tables/capacity-ladder}
\par\vspace{2pt}{\footnotesize $^{\dagger}$Earlier fits, with a cosine learning-rate schedule.\par}
\label{tab:capacity-ladder}
\end{table}
\begin{table}[t]
\centering
\caption{\textbf{The conditional pass, the early step bin and the text stream carry more of the gain than their counterparts.}
Gain when one part of the full $\mathbf{s}_{\mathrm{fit}}$ is added alone, in OCR reward and raw PickScore points.
SD3.5-M uses the steering vectors and test sets of \cref{tab:recovery-table}, with blocks counted from 0.}
\small
\setlength{\tabcolsep}{4pt}
\input{tables/part-ablation-probe}
\label{tab:part-ablation-probe}
\end{table}
\begin{table*}[t]
\centering
\caption{\textbf{Two of six video RL-tuned models gain on the reward we evaluate, and the steering vectors recover the gain of both.}
Top: each video RL-tuned model's gain on a reward it trained on, or on the rule score for RLVLM.
Bottom: $\mathbf{s}_{\mathrm{fit}}$ fitted to GenRL GRPO at step 1,500 and to VBVR-Pro RLVR, evaluated on 81-frame clips unless a row names 33 frames.}
\small
\setlength{\tabcolsep}{3.5pt}
\input{tables/video-detail}
\par\vspace{2pt}{\footnotesize $^{*}$RLVLM trains on a VLM judge, which agrees with the rule score on pass or fail for 75\% to 78\% of clips, so we score it with the rule score.\par}
\label{tab:video-detail}
\end{table*}
\begin{table}[t]
\centering
\caption{\textbf{The steering vectors solve most of the test pairs that the RL-tuned model fixes.}
Share of the prompt-seed pairs that the base model fails and the RL-tuned model solves that each SD3.5-M model also solves.
The last column is in percentage points.}
\small
\setlength{\tabcolsep}{4pt}
\input{tables/paired-success}
\label{tab:paired-success}
\end{table}
On the OCR and PickScore test sets, the restricted sets of \cref{fig:gain-localisation} gain $+0.342$ and $+1.500$ (conditional pass), $+0.315$ and $+1.343$ (text stream), $+0.315$ and $+1.325$ (image stream), $+0.310$ and $+1.161$ (early step bin), and $+0.308$ and $+1.150$ (all three).
\Cref{tab:recovery-details} gives the intervals behind each row of \cref{tab:recovery-table}, the Wan2.1 shares on each dimension of VideoAlign and on its overall score, and the out-of-domain split of VBVR-Pro-Bench, which holds task families outside the RL training tasks.
This split holds 250 instances of 50 task families, at four sampling seeds.
On Wan2.1, HPSv3 rates every frame against a fixed quality prompt, and HPSv3 percentile rates the top 30\% of frames against the clip's prompt.
\Cref{tab:hyperparameters} gives the training length of each row.
On the out-of-domain split, the steering vectors gain $+0.0397$ [$+0.0195$, $+0.0614$] and the VBVR-Pro RL-tuned model gains $+0.0474$ [$+0.0269$, $+0.0700$].
\Cref{tab:models-and-samplers} lists the checkpoint, update type, reward and sampler of the eight RL-tuned models of \cref{tab:recovery-table}.
The Wan2.1 RL-tuned model trained on two HPSv3 scores and on VideoAlign's motion quality and text alignment.
\Cref{tab:capacity-ladder} orders steering-vector layouts on SD3.5-M by their number of parameters.
Each fit has one training seed and is evaluated at one sampling seed.
Separate vectors for the conditional and unconditional passes add $+0.16$ [$+0.14$, $+0.18$] PickScore points at one step bin, whereas a second step bin adds $+0.03$.
A LoRA of rank 32 fitted by \cref{eq:fit}, with as many parameters as the RL-tuned model's LoRA, gains $+1.57$, against $+1.47$ for the steering vectors with two passes and two step bins and $+1.61$ for the RL-tuned model.
On 256 OCR prompts, a LoRA of rank 32 fitted for 1,500 updates gains $+0.331$, against $+0.324$ for steering vectors fitted for 1,500 updates and $+0.342$ for the RL-tuned model.
\Cref{tab:part-ablation-probe} gives the gain of each part of the steering vectors alone.
The SD3.5-M rows use one sampling seed.
The FLUX.1-dev rows use the PickScore-tuned model on 256 prompts at one sampling seed, with steering vectors fitted for 1,500 updates.
Each of its 38 single-stream blocks adds one vector to the image and text tokens alike.
Every FLUX.1-dev row keeps the double-stream vectors and adds the single-block vectors only to the listed tokens, except the last row, a separate fit with one vector per token type.
Each SD3.5-M row adds one part of a full fitted set and omits the rest, whereas \cref{fig:gain-localisation} fits a new set restricted to that part.
For a guidance pass or a step bin the two agree within 0.002 OCR and 0.006 PickScore points, since \cref{eq:fit} fits each pass and step bin independently, and for a stream they differ.
On SD3.5-M, the vectors of the conditional pass, of the early step bin and of the text stream each gain more than their counterparts, and the vectors of the unconditional pass alone lower both rewards.
Guidance weights the conditional velocity by 4.5 and the unconditional velocity by $-3.5$, so when the vectors of the two passes are aligned, dropping the unconditional-pass vectors also enlarges the net change.
In the single-stream blocks of FLUX.1-dev, the shared vector acts through the image and text tokens together, and separate vectors for the two token types gain $-0.017$ [$-0.041$, $+0.005$] PickScore points against the shared vector, from the same noise.
No single block group carries the gain of the steering vectors, and most of the gain lies outside the base model's top principal components.
On the OCR and PickScore test sets of \cref{tab:recovery-table}, the steering vectors gain $+0.325$ OCR and $+1.416$ PickScore.
Keeping only blocks 0 to 7, 8 to 15 or 16 to 23 gains $+0.167$, $+0.193$ and $+0.048$ on OCR and $+0.544$, $+0.746$ and $+0.318$ on PickScore.
The last eight blocks matter least: without them, the steering vectors gain $+0.323$ on OCR and $+1.154$ on PickScore, which is 99\% of the OCR gain and 81\% of the PickScore gain.
We also split the steering vector of each block into its projection onto the base model's top principal components and the residual, on 128 test prompts per reward with one sampling seed, for steering vectors fitted for 1,500 updates on OCR and 4,500 updates on PickScore.
Without the top-8 component, the steering vectors reach 106\% of the OCR gain and keep 79\% of the PickScore gain, and the top-8 component alone keeps 8\% and $-14$\%.
With the top 64 and 256 components, the residual reaches 107\% and 88\% of the OCR gain and 79\% and 73\% of the PickScore gain.
In the text stream, the top 8 principal components hold 64\% of the squared norm of the OCR steering vectors and 54\% of the PickScore ones, so the largest component of the text-stream vectors is not the one that carries the gain.
\Cref{tab:video-detail} lists the six video RL-tuned models we evaluated and the share of the RL gain that the steering vectors recover at each fitting setting.
GenRL GRPO trained on four rewards, and its rows give VideoAlign text alignment, the reward with the lowest interval.
SOLACE trains on its own denoising confidence and samples with CFG 5.0.
RAVEN is a causal, distilled Wan2.1-T2V-1.3B with four sampling steps, and CM-GRPO trains it on a weighted sum of five rewards.
Each Wan2.1 fit trains on 33-frame states for at most 1,500 updates, and Batch is the number of training states per update.
The text-context and key-input rows start from zero and add a constant to the whole cross-attention text context of each block or only to the input of its key projection.
The Wan2.1 rows use 64 VBench prompts and the RAVEN row 100 prompts.
The VBVR-Pro rows use 250 instances of the 50 RL training tasks other than the in-domain test instances of \cref{tab:recovery-table}.
Two of the six, GenRL GRPO and VBVR-Pro RLVR, gain on the reward we evaluate with intervals above zero, and \cref{tab:recovery-table} uses these two.
On Wan2.1, the steering vectors recover the HPSv3 gain at a batch of 16 and of 64 training states, and the batch of 64 raises the motion-quality share from 64\% to 108\%.
On these prompts, a constant on the whole cross-attention text context of each block recovers 95\% of the HPSv3 gain, and a constant on the key input alone recovers 1\%, since a shift shared by all text keys cancels in cross-attention and only the RMSNorm after the key projection leaves a small change (\cref{thm:attention}). On 128 of the test prompts of \cref{tab:recovery-table}, the two constants recover 108.4\% [96.5, 121.9] and $-8.0$\% [$-21.3$, 4.5].
On the OCR and PickScore test sets of \cref{tab:recovery-table}, $\mathbf{s}_{\mathrm{mean}}$ gains $+0.074$ OCR and $+0.952$ PickScore, and rescaled to the norms of $\mathbf{s}_{\mathrm{fit}}$ it gains $+0.201$ and $+1.231$.
Conversely, $\mathbf{s}_{\mathrm{fit}}$ rescaled to the norms of $\mathbf{s}_{\mathrm{mean}}$ gains $+0.157$ and $+1.123$, its component along $\mathbf{s}_{\mathrm{mean}}$ alone gains $+0.078$ and $+0.920$, and its component orthogonal to $\mathbf{s}_{\mathrm{mean}}$ alone gains $+0.312$ and $+0.542$.

\Cref{tab:paired-success} tests whether the steering vectors solve the same prompts and starting noises as the RL-tuned model.
A pair is one test prompt and one sampling seed, on the GenEval test set of 553 prompts with four sampling seeds and the OCR test set of 1,018 prompts with three sampling seeds.
A pair counts as solved when GenEval's check is correct or when the text read from the image matches the quoted text exactly.
The base model solves 66.1\% of the GenEval pairs and 21.4\% of the OCR pairs, the RL-tuned models 94.0\% and 62.9\%, and the steering vectors 93.0\% and 60.2\%.
These rates average over pairs, whereas the GenEval score of \cref{tab:recovery-table} averages over the six tasks, so the GenEval gains of the two differ.
Of the 641 GenEval pairs that the base model fails and the GenEval-tuned model solves, the steering vectors also solve 91.4\%, with no detected difference from the 93.3\% that the GenEval-tuned model itself solves at the next sampling seed of the same prompt.
Of the 1,382 such OCR pairs, the steering vectors solve 75.3\%, which exceeds the OCR-tuned model's own 71.6\% at the next sampling seed by $+3.8$ [$+0.7$, $+6.8$] points.
The mean-difference steering vectors solve 30.1\% of the GenEval pairs that the GenEval-tuned model fixes.
A same-size LoRA fitted by \cref{eq:fit} solves 90.6\% of these GenEval pairs.
At sampling seed 0, the one seed with OCR scores for this LoRA, it solves 75.4\% of the 460 OCR pairs that the OCR-tuned model fixes, and the steering vectors solve 77.6\%.
On both sets, no difference between the same-size LoRA and the steering vectors is detected.
On the pairs that both the base model and the RL-tuned model fail, the steering vectors solve 33.3\% of 108 on GenEval and 27.3\% of 1,017 on OCR.

\begin{figure}[t]
\centering
\includegraphics{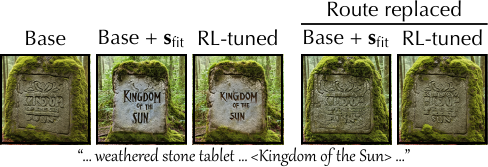}
\caption{\textbf{Replacing what image tokens read from text removes the rendered text.}
An OCR test prompt. In the last two images, image tokens read the base model's text keys and values.}
\label{fig:attention-route-b}
\end{figure}

\paragraph{Text keys and values from the steered base model.}
Finding 2 of \cref{sec:ana_credit} removes most of the OCR-tuned model's gain by replacing the text keys and values that its image queries read with those of the base model.
If the steering vectors carry the change that RL makes to these keys and values, computing them with the steered base model should restore the gain.
We therefore repeat the replacement on the OCR-tuned model, with the text keys and values computed by the base model plus the OCR steering vectors on the same noisy state, on the OCR test set of 1,018 prompts at one sampling seed.
The OCR-tuned model then keeps 95.2\% [91.9, 98.6] of its gain, against 26.4\% [20.7, 31.6] with the keys and values of the unsteered base model.
With the PickScore steering vectors in place of the OCR steering vectors, it keeps 44.0\% [38.2, 49.5], so the restored gain is specific to the reward that the steering vectors were fitted for.

\paragraph{More samples.}
\Cref{fig:sample-gallery} shows the base model, the steered base model and each SD3.5-M RL-tuned model of \cref{tab:recovery-table} on test prompts.
\begin{figure*}[p]
\centering
\includegraphics[width=\textwidth]{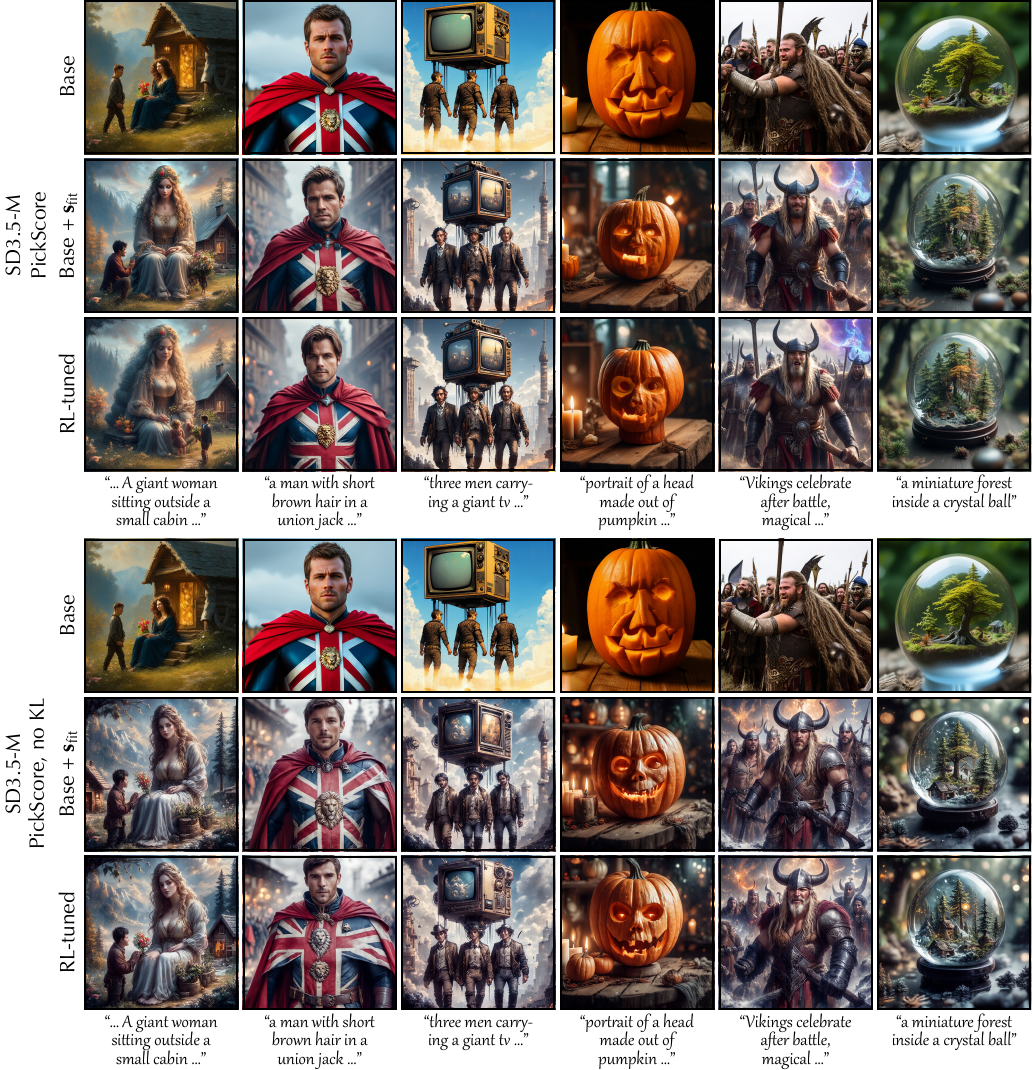}
\caption{\textbf{The steered base model mostly draws what the RL-tuned model draws.}
Each strip shows one SD3.5-M RL-tuned model, and within a strip each column shares one test prompt and starting noise.
Both PickScore strips show the same prompts and base images.}
\label{fig:sample-gallery}
\end{figure*}
\begin{figure*}[p]
\ContinuedFloat
\centering
\includegraphics[width=\textwidth]{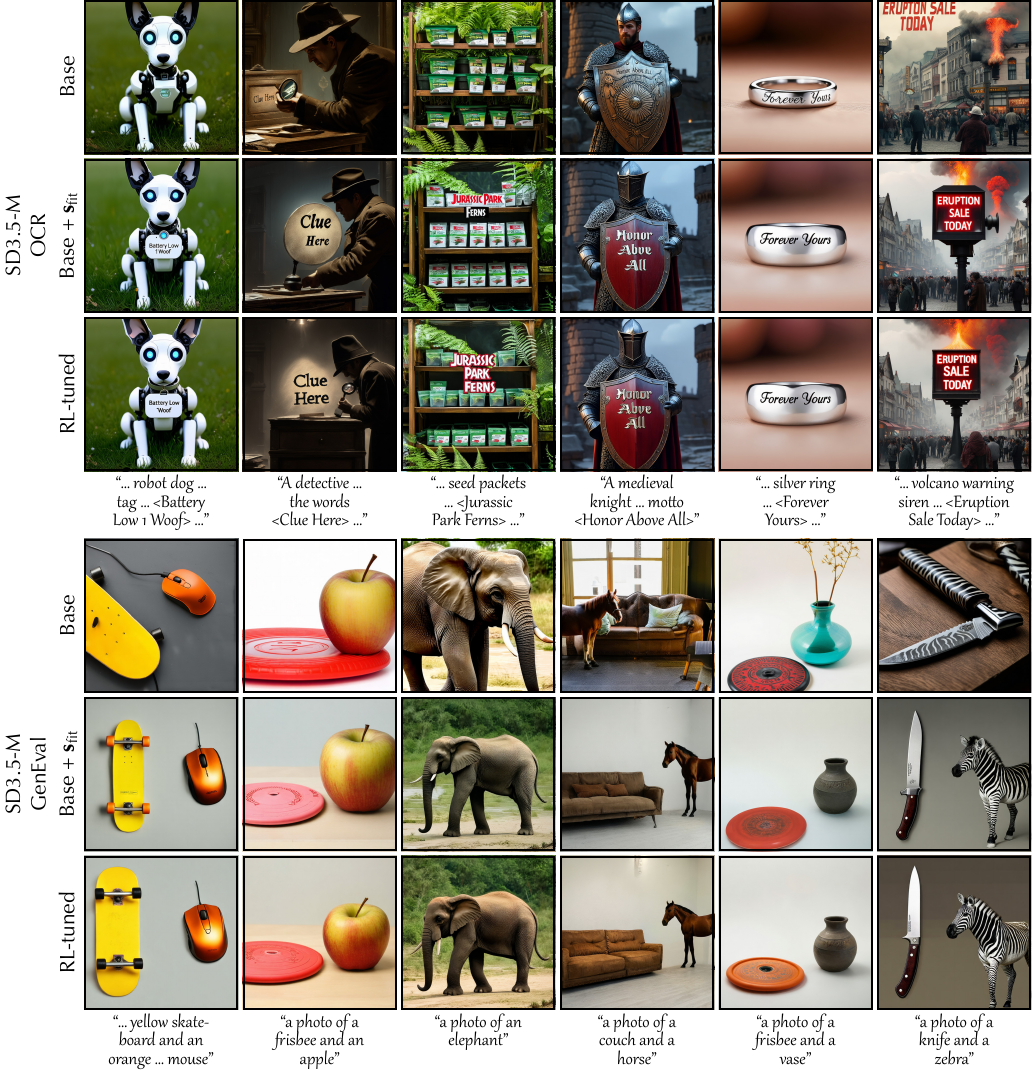}
\caption{\textbf{(continued)} The OCR-tuned and GenEval-tuned models.}
\end{figure*}

\paragraph{Full prompts.}
Where a figure shortens or omits a prompt, the list below gives the full prompt, in the order of the figure.
\begin{itemize}[nosep,leftmargin=*]
\item \Cref{fig:teaser}, OCR: \emph{An astronaut standing on the gray surface of the moon with the blue Earth rising behind, holding up a white sign that reads ``All That RL for a Constant Shift?'' in bold black letters.}
\item \Cref{fig:teaser}, OCR: \emph{A realistic photograph of a modern city street, with a large, vibrant sign that says ``Deep Learning'' illuminated in neon lights, set against a backdrop of bustling city life and futuristic architecture.}
\item \Cref{fig:shared-shift}, first prompt: \emph{A vibrant autumn leaf collage titled ``Fall Colors'', featuring a harmonious mix of red, orange, and yellow leaves arranged in a natural, artistic pattern against a subtle white background.}
\item \Cref{fig:shared-shift}, second prompt: \emph{A vibrant, realistic photograph of a DJ turntable with a colorful sticker on its side declaring ``Spin Masters'', set against a backdrop of a bustling, neon-lit nightclub.}
\item \Cref{fig:image-examples}, OCR-tuned model: \emph{A vibrant food truck parked on a bustling city street, its side panel airbrushed with the bold slogan ``Tacos Save Lives'', surrounded by eager customers and the aroma of sizzling tacos.}
\item \Cref{fig:image-examples}, PickScore-tuned model: \emph{enigmatic black square building on top of a purple hill, smoke stacks}
\item \Cref{fig:image-examples}, PickScore-tuned model without KL: \emph{A photo of an astronaut riding a horse in the forest. There is a river in front of them with water lilies.}
\item \Cref{fig:image-examples}, FLUX.1-dev HPSv2.1-tuned model: \emph{Petite woman with Short brown hair wearing t-shirt and shorts doing yoga}
\item \Cref{fig:video-frames}, Wan2.1-1.3B: \emph{A cute happy Corgi playing in park, sunset, tilt down}
\item \Cref{fig:video-frames}, Wan2.2-5B: \emph{The scene has two separated areas: a top SEQUENCE area and a bottom CHOICES area. In the SEQUENCE area, the shapes are the same shape and the same color, and their sizes strictly increase from left to right. First identify the constant size step between consecutive sequence shapes, then select the one correct option (out of 4) in the CHOICES area that continues the same shape, color, and size-increase pattern. Circle the correct option with a red circle and show the full process step by step. Only modify what the task requires; keep the background and any other element unchanged.}
\item \Cref{fig:video-frames}, Wan2.2-5B: \emph{The scene has two separated areas: a top SEQUENCE area and a bottom CHOICES area. In the SEQUENCE area, the shapes are the same shape and the same color, and their sizes strictly alternate between LARGE and SMALL from left to right. First observe the size-alternation pattern and determine whether the next item should be LARGE or SMALL, then select the one correct option (out of 4) in the CHOICES area that continues the same shape, color, and large/small alternation pattern. Circle the correct option with a red circle and show the full process step by step. Only modify what the task requires; keep the background and any other element unchanged.}
\item \Cref{fig:rl-steering-examples}, OCR: \emph{A vintage library stamp, prominently marked ``Forbidden Knowledge'', is centered on an aged, parchment-like paper. The stamp features intricate, almost mystical designs around the text, suggesting a wealth of secret wisdom. The paper shows signs of wear, with faded edges and a few small tears, enhancing the sense of antiquity and mystery.}
\item \Cref{fig:rl-steering-examples}, OCR: \emph{A bustling city street features a bicycle rental kiosk with a prominent sign that reads ``Rent A Bike Here'', surrounded by a variety of colorful bikes and happy cyclists. The scene is vibrant and lively, capturing the essence of urban cycling culture.}
\item \Cref{fig:rl-steering-examples}, PickScore: \emph{key frame anime still of a pigeon in a well tailored suit getting a cup of coffee in a cafe in the morning}
\item \Cref{fig:attention-route-b}: \emph{An ancient, weathered stone tablet, partially covered in moss, with the intricate inscription ``Kingdom of the Sun'' prominently carved into its surface, set against a backdrop of dense, overgrown forest.}
\item \Cref{fig:transfer-examples}, PickScore, left: \emph{Beautiful image presenting that AGI is coming, digital concept art, sci-fi, cyberpunk, superintelligence supercomputer, futuristic, monumental AI singularity oracle, breathtaking, bottom view, intricate details}
\item \Cref{fig:transfer-examples}, OCR, left: \emph{A close-up photograph of an engraved silver ring with the inscription ``Forever Yours'' delicately etched into its surface, set against a soft, blurred background of romantic, warm tones.}
\item \Cref{fig:transfer-examples}, OCR, right: \emph{A ``party'' reminder poster, brightly colored and eye-catching, is stuck on the side window of a bustling city bus, partially obscured by the reflection of passengers chatting and laughing inside, under the glow of evening streetlights.}
\item \Cref{fig:side-effect-examples}: \emph{nature vs human, surreal, UHD, HDR, 8K, hyper details, rich colors, photograph}
\item \Cref{fig:sample-gallery}, PickScore: \emph{Fantasy, Masterpiece, A giant woman sitting outside a small cabin with her tiny lover coming out to give her flowers.}
\item \Cref{fig:sample-gallery}, PickScore: \emph{a man with short brown hair in a union jack outfit with long cape, emblem of a lion on the front, directed by Zack Snyder, cinematic high res, 8k, award winning}
\item \Cref{fig:sample-gallery}, PickScore: \emph{three men carrying a giant tv above their head, steampunk and anime style}
\item \Cref{fig:sample-gallery}, PickScore: \emph{portrait of a head made out of pumpkin, on a wooden table, candle-lit from behind}
\item \Cref{fig:sample-gallery}, PickScore: \emph{Vikings celebrate after battle, magical, mystical, fantastical}
\item \Cref{fig:sample-gallery}, OCR: \emph{A realistic photograph of a futuristic robot dog sitting on a grassy lawn, with a metallic collar featuring a small, circular tag that reads ``Battery Low 1 Woof'' in clear, bold text. The dog's eyes glow softly, and its body shows subtle mechanical details.}
\item \Cref{fig:sample-gallery}, OCR: \emph{A detective in a trench coat and hat examines a crime scene, his magnifying glass revealing the words ``Clue Here'' etched into a dusty, old wooden desk. The scene is dimly lit, with shadows cast by a single, flickering overhead light.}
\item \Cref{fig:sample-gallery}, OCR: \emph{A plant nursery shelf filled with seed packets, prominently labeled ``Jurassic Park Ferns'', surrounded by lush, prehistoric-looking ferns and other ancient plants, with a subtle mist adding an ethereal atmosphere.}
\item \Cref{fig:sample-gallery}, OCR: \emph{A medieval knight holds a shield, its surface intricately engraved with the motto ``Honor Above All'', standing against a backdrop of an ancient stone castle under a twilight sky.}
\item \Cref{fig:sample-gallery}, OCR: \emph{A close-up photograph of an engraved silver ring with the inscription ``Forever Yours'' delicately etched into its surface, set against a soft, blurred background of romantic, warm tones.}
\item \Cref{fig:sample-gallery}, OCR: \emph{A dramatic scene with a volcano warning siren flashing the message ``Eruption Sale Today'' in a bustling town square, where panicked citizens and curious onlookers gather, blending realism with surrealism.}
\item \Cref{fig:sample-gallery}, GenEval: \emph{a photo of a yellow skateboard and an orange computer mouse}
\end{itemize}

%% file: tables/hyperparameters.tex
\begin{tabular}{@{}l*{8}{c}@{}}
\toprule
 & \multicolumn{4}{c}{SD3.5-M} & \multicolumn{2}{c}{FLUX.1-dev} & Wan2.1-1.3B & Wan2.2-5B \\
\cmidrule(lr){2-5}\cmidrule(lr){6-7}\cmidrule(lr){8-8}\cmidrule(l){9-9}
 & PickScore & \begin{tabular}[b]{@{}c@{}}PickScore,\\ no KL\end{tabular} & OCR & GenEval & PickScore & HPSv2.1 & HPSv3 & VBVR-Pro \\
\midrule
Learning rate & $3{\times}10^{-3}$ & $10^{-2}$ & $3{\times}10^{-3}$ & $10^{-2}$ & $3{\times}10^{-3}$ & $3{\times}10^{-3}$ & $3{\times}10^{-3}$ & $3{\times}10^{-3}$ \\
LR decay from & 3,750 & 3,750 & none & none & 1,250 & 1,250 & 1,250 & 1,250 \\
States per update & 64 & 64 & 64 & 64 & 64 & 32 & 64 & 64 \\
Training prompts & 10,000 & 10,000 & 10,000 & 10,000 & 2,048 & 1,024 & 512 & 512 \\
Training states & new & new & new & new & 8 fixed & 16 fixed & 8 fixed & 8 fixed \\
Training updates & 4,500 & 4,500 & 3,000 & 3,000 & 1,500 & 1,500 & 1,500 & 1,500 \\
\midrule
H100 GPU-hours & 8.9 & 8.9 & 6.9 & 6.9 & 5.6 & 16.8 & 69.7 & 22.4 \\
\bottomrule
\end{tabular}

%% file: tables/shift-layouts.tex
\begin{tabular}{@{}lrrrr@{}}
\toprule
Base model & Block outputs & Width & Passes & Parameters \\
\midrule
SD3.5-M & 47 & 1,536 & 2 & 288,768 \\
FLUX.1-dev & 76 & 3,072 & 1 & 466,944 \\
Wan2.1-1.3B & 30 & 1,536 & 2 & 184,320 \\
Wan2.2-5B & 30 & 3,072 & 1 & 184,320 \\
\bottomrule
\end{tabular}

%% file: tables/step-bins.tex
\begin{tabular}{@{}lllr@{}}
\toprule
Base model & RL method & Reward & Kept (\%) \\
\midrule
SD3.5-M & Flow-GRPO & PickScore & 85 \\
 & Flow-GRPO, no KL & PickScore & 94 \\
 & Flow-GRPO & OCR & 83 \\
 & Flow-GRPO & GenEval & 88 \\
\midrule
FLUX.1-dev & Flow-GRPO & PickScore & 94 \\
 & DanceGRPO & HPSv2.1 & 86 \\
\bottomrule
\end{tabular}

%% file: tables/controls-table.tex
\begin{tabular}{@{}lllrrrrr@{}}
\toprule
 & & & & \multicolumn{3}{c}{Share (\%)} & \\
\cmidrule(lr){5-7}
Base & RL method & Reward & RL-tuned gain & $\mathbf{s}_{\mathrm{fit}}$ & LoRA & Rank-1 RL LoRA & $\mathbf{s}_{\mathrm{fit}}$ minus LoRA \\
\midrule
SD3.5-M & Flow-GRPO & PickScore & +1.560 & 90.8 & 90.8 & 43.1 & 0.000 [$-$0.013, +0.013] \\
 & Flow-GRPO & OCR & +0.344 & 94.4 & 97.0 & 62.8 & $-$0.009 [$-$0.022, +0.003] \\
 & Flow-GRPO & GenEval & +0.271 & 95.9 & 93.4 & -- & +0.007 [$-$0.004, +0.017] \\
\midrule
FLUX.1-dev & Flow-GRPO & PickScore & +1.346 & 92.1 & 89.5 & -- & +0.036 [$-$0.005, +0.079] \\
 & DanceGRPO & HPSv2.1 & +0.055 & 99.2 & 97.4 & -- & +0.001 [0.000, +0.002] \\
\midrule
Wan2.1-1.3B & GenRL GRPO & HPSv3 & +1.309 & 111.9 & 69.7 & -- & +0.552 [+0.465, +0.641] \\
\midrule
Wan2.2-5B & VBVR-Pro RLVR & Rule score & +0.080 & 89.8 & 82.1 & -- & +0.006 [$-$0.007, +0.020] \\
\bottomrule
\end{tabular}

%% file: tables/other-interventions.tex
\begin{tabular}{@{}lrrr@{}}
\toprule
Intervention & Parameters & PickScore & OCR \\
\midrule
Steering vectors $\mathbf{s}_{\mathrm{fit}}$ & 288,768 & +1.43 & +0.324 \\
Conditioning offset, fitted & 6,144 & +0.42 & +0.032 \\
Gate scales, fitted & 432 & +0.26 & +0.027 \\
Prompt suffix & 0 & +0.06 & +0.011 \\
CFG scale 6 & 0 & +0.00 & +0.032 \\
CFG scale 7.5 & 0 & $-$0.05 & +0.040 \\
\bottomrule
\end{tabular}

%% file: tables/user-study.tex
\setlength{\tabcolsep}{3.5pt}
\begin{tabular}{@{}llcccccc@{}}
\toprule
 & & \multicolumn{3}{c}{vs Base} & \multicolumn{3}{c}{vs RL-tuned} \\
\cmidrule(lr){3-5}\cmidrule(l){6-8}
Reward & Vectors & PF & Aes & All & PF & Aes & All \\
\midrule
PickScore & $\mathbf{s}_{\mathrm{fit}}$ & 65 & 67 & 64 & 49 & 53 & 51 \\
 & $\mathbf{s}_{\mathrm{RL}}$ & 68 & 67 & 62 & 51 & 48 & 52 \\
\addlinespace
OCR & $\mathbf{s}_{\mathrm{fit}}$ & 62 & 67 & 60 & 49 & 47 & 50 \\
 & $\mathbf{s}_{\mathrm{RL}}$ & 63 & 60 & 65 & 52 & 50 & 52 \\
\bottomrule
\end{tabular}

%% file: tables/recovery-details.tex
\begin{tabular}{@{}lllll@{}}
\toprule
 & & & \multicolumn{2}{c}{Share of the RL gain (\%)} \\
\cmidrule(l){4-5}
Base & RL-tuned model or score & RL-tuned gain & $\mathbf{s}_{\mathrm{mean}}$ & $\mathbf{s}_{\mathrm{fit}}$ \\
\midrule
SD3.5-M & PickScore & +1.560 [+1.521, +1.599] & 61.0 [59.7, 62.3] & 90.8 [89.8, 91.7] \\
 & PickScore, no KL & +1.693 [+1.648, +1.734] & 70.8 [69.4, 72.2] & 95.2 [94.4, 96.1] \\
 & OCR & +0.344 [+0.319, +0.369] & 21.5 [15.4, 26.9] & 94.4 [90.8, 97.9] \\
 & GenEval & +0.271 [+0.250, +0.293] & 12.2 [6.5, 17.5] & 95.9 [92.0, 99.7] \\
\midrule
FLUX.1-dev & Flow-GRPO, PickScore & +1.522 [+1.481, +1.561] & 64.7 [62.8, 66.6] & 90.0 [88.9, 91.1] \\
 & DanceGRPO, HPSv2.1 & +0.0536 [+0.0523, +0.0550] & 79.0 [77.6, 80.5] & 96.3 [95.5, 97.2] \\
\midrule
Wan2.1-1.3B & HPSv3 & +1.309 [+1.198, +1.424] & 59.3 [54.8, 63.9] & 111.9 [106.5, 118.0] \\
 & HPSv3 percentile & +1.615 [+1.424, +1.828] & 60.2 [55.1, 65.2] & 105.6 [99.3, 112.1] \\
 & VideoAlign MQ & +0.219 [+0.166, +0.276] & 38.1 [20.0, 53.5] & 87.5 [68.5, 106.7] \\
 & VideoAlign TA & +0.113 [+0.074, +0.151] & 60.7 [36.1, 88.8] & 79.4 [51.1, 109.1] \\
 & VideoAlign VQ$^{*}$ & +0.296 [+0.255, +0.339] & 49.4 [40.5, 58.3] & 111.7 [101.5, 123.3] \\
 & VideoAlign overall & +0.628 [+0.538, +0.725] & 47.5 [37.8, 56.7] & 97.4 [86.6, 109.2] \\
\midrule
Wan2.2-5B & In-domain instances & +0.0800 [+0.0617, +0.0991] & 60.8 [45.3, 77.3] & 89.8 [76.2, 105.6] \\
 & Out-of-domain split & +0.0474 [+0.0269, +0.0700] & 41.4 [13.2, 72.5] & 83.9 [52.7, 123.7] \\
\bottomrule
\end{tabular}

%% file: tables/models-and-samplers.tex
\begin{tabular}{@{}llllrrl@{}}
\toprule
Base & RL method & Update & Reward & Steps & Guidance & Resolution \\
\midrule
SD3.5-M & Flow-GRPO & LoRA, $r=32$ & PickScore & 40 & 4.5 & 512 \\
 & \multicolumn{6}{@{}l}{\footnotesize \texttt{jieliu/SD3.5M-FlowGRPO-PickScore}} \\[1pt]
 & Flow-GRPO, no KL & LoRA, $r=32$ & PickScore & 40 & 4.5 & 512 \\
 & \multicolumn{6}{@{}l}{\footnotesize \texttt{jieliu/SD3.5M-FlowGRPO-PickScore-without-KL}} \\[1pt]
 & Flow-GRPO & LoRA, $r=32$ & OCR & 40 & 4.5 & 512 \\
 & \multicolumn{6}{@{}l}{\footnotesize \texttt{jieliu/SD3.5M-FlowGRPO-Text}} \\[1pt]
 & Flow-GRPO & LoRA, $r=32$ & GenEval & 40 & 4.5 & 512 \\
 & \multicolumn{6}{@{}l}{\footnotesize \texttt{jieliu/SD3.5M-FlowGRPO-GenEval}} \\[1pt]
\midrule
FLUX.1-dev & Flow-GRPO & LoRA, $r=64$ & PickScore & 28 & 3.5 & 512 \\
 & \multicolumn{6}{@{}l}{\footnotesize \texttt{shi-labs/FLUX.1-dev-ind-expert-PickScore}} \\[1pt]
 & DanceGRPO & Full & HPSv2.1 & 50 & 3.5 & 1024 \\
 & \multicolumn{6}{@{}l}{\footnotesize \texttt{xzyhku/flux\_hpsv2.1\_dancegrpo (checkpoint-300-0)}} \\[1pt]
\midrule
Wan2.1-1.3B & GenRL GRPO & LoRA, $r=128$ & HPSv3, VideoAlign & 50 & 4.5 & $480{\times}832$, 81 frames \\
 & \multicolumn{6}{@{}l}{\footnotesize \texttt{lightx2v/Wan2.1-T2V-1.3B-longcat-step1500}} \\[1pt]
\midrule
Wan2.2-5B & VBVR-Pro RLVR & Full & Rule score & 30 & 1 & $512{\times}512$, 81 frames \\
 & \multicolumn{6}{@{}l}{\footnotesize \texttt{Video-Reason/VBVR-Pro-Wan2.2-TI2V-5B-RLVR @ 019fa77}} \\[1pt]
\bottomrule
\end{tabular}

%% file: tables/capacity-ladder.tex
\begin{tabular}{@{}rlrrrrr@{}}
\toprule
Passes & Streams & Bins & Parameters & Updates & LR & Gain \\
\midrule
\multicolumn{7}{@{}l}{Mean-difference steering vectors $\mathbf{s}_{\mathrm{mean}}$} \\
1 & Both & 8 & 577,536 & -- & -- & +0.62 \\
2 & Both & 8 & 1,155,072 & -- & -- & +0.99 \\
\midrule
\multicolumn{7}{@{}l}{Fitted steering vectors $\mathbf{s}_{\mathrm{fit}}$} \\
1 & Image & 1 & 36,864 & 1,800 & $10^{-2}$ & +1.11 \\
1 & Both & 1 & 72,192 & 1,800 & $10^{-2}$ & +1.30 \\
1 & Both & 2 & 144,384 & 3,600 & $10^{-2}$ & +1.33 \\
2 & Both & 1 & 144,384 & 1,800 & $10^{-2}$ & +1.46 \\
1 & Both & 4 & 288,768 & 3,000 & $3{\times}10^{-3}$ & +1.29 \\
2 & Both & 2 & 288,768 & 4,500 & $3{\times}10^{-3}$ & +1.47 \\
1 & Both & 8 & 577,536 & 1,500$^{\dagger}$ & $3{\times}10^{-3}$ & +1.20 \\
2 & Both & 4 & 577,536 & 3,600 & $3{\times}10^{-3}$ & +1.45 \\
\midrule
\multicolumn{3}{@{}l}{LoRA, rank 32, fitted} & 18.8M & 1,500$^{\dagger}$ & $3{\times}10^{-4}$ & +1.57 \\
\multicolumn{3}{@{}l}{RL-tuned model} & 18.8M & -- & -- & +1.61 \\
\bottomrule
\end{tabular}

%% file: tables/part-ablation-probe.tex
\begin{tabular}{@{}lrr@{}}
\toprule
Steering vectors kept & OCR & PickScore \\
\midrule
\multicolumn{3}{@{}l}{SD3.5-M, test prompts} \\
\quad RL-tuned model & +0.344 & +1.560 \\
\quad All steering vectors & +0.325 & +1.416 \\
\quad Conditional pass $\mathrm{c}$ only & +0.344 & +1.500 \\
\quad Unconditional pass $\varnothing$ only & $-$0.046 & $-$1.335 \\
\quad Step bin $t \ge \tau$ only & +0.309 & +1.155 \\
\quad Step bin $t < \tau$ only & +0.044 & +0.244 \\
\quad Text stream only & +0.254 & +1.310 \\
\quad Image stream only & +0.130 & +0.016 \\
\quad Blocks 0--7 only & +0.167 & +0.544 \\
\quad Blocks 8--15 only & +0.193 & +0.746 \\
\quad Blocks 16--23 only & +0.048 & +0.318 \\
\quad Blocks 0--15 only & +0.323 & +1.154 \\
\midrule
\multicolumn{3}{@{}l}{FLUX.1-dev single blocks} \\
\quad RL-tuned model & -- & +1.346 \\
\quad Image and text tokens & -- & +1.240 \\
\quad Image tokens only & -- & +1.053 \\
\quad Text tokens only & -- & +0.615 \\
\quad No single-block vectors & -- & +0.923 \\
\quad Separate text and image vectors, fitted & -- & +1.223 \\
\bottomrule
\end{tabular}

%% file: tables/video-detail.tex
\begin{tabular}{@{}lllrll@{}}
\toprule
Base & RL-tuned model or fit & Reward & Prompts $\times$ seeds & Gain over base & Share (\%) \\
\midrule
\multicolumn{6}{@{}l}{RL-tuned models, gain over the base model} \\
Wan2.1-1.3B & GenRL GRPO, RL step 500 & VideoAlign TA & 64 $\times$ 2 & +0.058 [$-$0.005, +0.124] & -- \\
 & GenRL GRPO, RL step 1,000 & VideoAlign TA & 64 $\times$ 2 & +0.135 [+0.075, +0.198] & -- \\
 & GenRL GRPO, RL step 1,500 & VideoAlign TA & 64 $\times$ 2 & +0.091 [+0.015, +0.168] & -- \\
 & SOLACE & Self-confidence & 64 $\times$ 2 & $-$1.115 [$-$1.211, $-$1.021] & -- \\
\addlinespace[2pt]
RAVEN & CM-GRPO & Sum of five rewards & 100 $\times$ 2 & +0.128 [$-$0.001, +0.262] & -- \\
\addlinespace[2pt]
Wan2.2-5B & VBVR-Pro RLVR & Rule score & 250 $\times$ 4 & +0.0561 [+0.0404, +0.0722] & -- \\
 & VBVR-Pro SDE-RLVR & Rule score & 250 $\times$ 4 & +0.0037 [$-$0.0145, +0.0214] & -- \\
 & VBVR-Pro RLVLM & Rule score$^{*}$ & 250 $\times$ 4 & +0.0125 [$-$0.0001, +0.0250] & -- \\
\midrule
\multicolumn{6}{@{}l}{Steering vectors $\mathbf{s}_{\mathrm{fit}}$, by fitting and evaluation setting} \\
Wan2.1-1.3B & Batch 16, 81 frames & HPSv3 & 64 $\times$ 2 & +1.487 [+1.278, +1.705] & 109 [94, 127] \\
 & Batch 16, 81 frames & VideoAlign MQ & 64 $\times$ 2 & +0.141 [+0.007, +0.277] & 64 [4, 112] \\
 & Batch 16, 33 frames & HPSv3 & 64 $\times$ 2 & +1.308 [+1.070, +1.556] & 106 [89, 129] \\
 & Batch 16, 33 frames & VideoAlign MQ & 64 $\times$ 2 & +0.016 [$-$0.057, +0.085] & 23 [$-$225, 129] \\
 & Batch 64, 81 frames & HPSv3 & 64 $\times$ 2 & +1.455 [+1.228, +1.701] & 107 [97, 117] \\
 & Batch 64, 81 frames & VideoAlign MQ & 64 $\times$ 2 & +0.237 [+0.117, +0.363] & 108 [68, 163] \\
 & Text context, batch 64 & HPSv3 & 64 $\times$ 2 & +1.297 [+1.087, +1.531] & 95 \\
 & Key input, batch 64 & HPSv3 & 64 $\times$ 2 & +0.014 [$-$0.226, +0.267] & 1 \\
\addlinespace[2pt]
Wan2.2-5B & Batch 64 & Rule score & 250 $\times$ 4 & +0.0458 [+0.0293, +0.0624] & 82 [62, 100] \\
\bottomrule
\end{tabular}

%% file: tables/paired-success.tex
\begin{tabular}{@{}lll@{}}
\toprule
Model & Share (\%) & $\mathbf{s}_{\mathrm{fit}}$ minus model \\
\midrule
\multicolumn{3}{@{}l}{GenEval test, seeds 0 to 3: 641 fixed, 108 both fail} \\
$\mathbf{s}_{\mathrm{fit}}$ & 91.4 [88.8, 93.9] & -- \\
$\mathbf{s}_{\mathrm{mean}}$ & 30.1 [26.0, 34.2] & +61.3 [+56.8, +65.8] \\
Same-size LoRA & 90.6 [87.9, 93.0] & +0.8 [$-$1.7, +3.3] \\
RL-tuned, next seed & 93.3 [91.1, 95.4] & $-$1.9 [$-$4.9, +1.0] \\
Both fail: $\mathbf{s}_{\mathrm{fit}}$ & 33.3 [22.8, 45.5] & -- \\
\midrule
\multicolumn{3}{@{}l}{OCR test, seeds 0 to 2: 1,382 fixed, 1,017 both fail} \\
$\mathbf{s}_{\mathrm{fit}}$ & 75.3 [73.0, 77.6] & -- \\
RL-tuned, next seed & 71.6 [69.1, 73.8] & +3.8 [+0.7, +6.8] \\
Both fail: $\mathbf{s}_{\mathrm{fit}}$ & 27.3 [24.5, 30.3] & -- \\
\midrule
\multicolumn{3}{@{}l}{OCR test, seed 0: 460 fixed} \\
$\mathbf{s}_{\mathrm{fit}}$ & 77.6 [73.9, 81.4] & -- \\
Same-size LoRA & 75.4 [71.6, 79.2] & +2.2 [$-$2.5, +7.1] \\
\bottomrule
\end{tabular}